\documentclass[a4paper,fleqn]{cas-sc}

\usepackage[numbers,sort&compress]{natbib}
\usepackage{amsmath,amssymb,amsfonts,bm}
\usepackage{mathtools}
\usepackage{algorithmic}
\usepackage{algorithm}
\usepackage{array}
\usepackage[labelfont=bf]{caption}
\usepackage{subcaption}
\renewcommand{\figurename}{Fig.}

\makeatletter
\renewcommand{\fnum@figure}{\figurename~\thefigure.\@gobble}
\makeatother

\begin{document}
\let\WriteBookmarks\relax
\def\floatpagepagefraction{1}
\def\textpagefraction{.001}

\shorttitle{Unscented KalmanNet with calibrated posterior covariance}

\shortauthors{M. Ko and A. Shafieezadeh}

\title [mode = title]{Unscented KalmanNet: Structure-Preserving Deep Learning with Calibrated Posterior Uncertainty under Incomplete Physics and Unknown Noise}                      

\author[1]{Minhyeok Ko}[type=editor,
                        orcid=0009-0003-9515-1687]
\cormark[1]
\ead{mko@uttyler.edu}
\credit{Conceptualization, Methodology, Software, Investigation, Validation, Visualization, Writing - original draft}
\affiliation[1]{organization={Department of Civil and Construction
                              Engineering and Management,
                              The University of Texas at Tyler},
                city={Tyler},
                state={TX},
                postcode={75799},
                country={USA}}

\author[2]{Abdollah Shafieezadeh}[orcid=0000-0001-6768-8522]
\ead{shafieezadeh.1@osu.edu}
\credit{Conceptualization, Methodology, Writing - review \& editing}
\affiliation[2]{organization={Department of Civil, Environmental, and Geodetic Engineering, The Ohio State University},
                city={Columbus},
                state={OH},
                postcode={43210},
                country={USA}}

\cortext[cor1]{Corresponding author}



\begin{abstract}
Nonlinear state estimation requires sequentially fusing model-based predictions with noisy measurements. Under imperfect dynamics and unknown, time-varying noise statistics, this fusion can degrade in both accuracy and statistical consistency. Existing learning-aided filters largely treat accuracy and uncertainty estimation separately, limitating their ability to correct model-mismatch-induced bias while retaining an explicit, calibrated posterior covariance. This paper introduces Unscented KalmanNet (UKN), a new model-based deep learning architecture that extends the Unscented Kalman Filter (UKF) with dedicated learned mechanisms for these two principal sources of filtering error, while preserving explicit posterior covariance propagation. NoiseNet learns time-varying process and measurement covariances as bounded multiplicative corrections to baseline covariances, guaranteeing positive definiteness, while GainNet learns a bounded residual correction to the analytical UKF gain to compensate for model-mismatch-induced bias. A new calibration-aware training objective is introduced that couples state estimation error with posterior covariance and innovation consistency terms through adaptive weighting, jointly optimizing accuracy and calibration. UKN is benchmarked against UKF, KalmanNet, and Bayesian KalmanNet on three synthetic systems and real-flight data from UZH-FPV. It achieves the lowest overall state-estimation error in all four examples and reduces RMSE by 26.4-49.7\% compared with UKF in the synthetic cases. Leave-one-sequence-out cross-validation over 11 flights shows 22.4\% and 34.3\% reductions in mean position and velocity RMSE, respectively. UKN also yields the lowest fold-to-fold variability, with dimension-normalized NEES and empirical coverage closest to nominal values among the covariance-reporting filters. These results demonstrate that structured learned adaptation can substantially improve estimation accuracy while retaining calibrated uncertainty.
\end{abstract}


\begin{highlights}
\item UKN combines model-based filtering with learned noise and gain corrections
\item NoiseNet predicts positive-definite time-varying process and measurement covariances
\item GainNet applies a bounded residual correction to the analytical UKF gain
\item Calibration-aware training jointly optimizes accuracy and covariance calibration
\item UKN achieves the lowest estimation error and best covariance calibration
\end{highlights}


\maketitle

\section{Introduction}
State estimation for nonlinear dynamical systems underlies a wide range of applications, including navigation \cite{rigatos2012nonlinear}, target tracking \cite{barshalom2001estimation}, robotics \cite{thrun2005probabilistic}, and structural health monitoring \cite{chatzi2009unscented, wu2007application}. From an information-processing perspective, recursive state estimation combines prior dynamic knowledge with incoming measurements to continuously update both the state estimate and its associated uncertainty. The recursive Bayesian filters of the Kalman family remain the dominant tools for this task \cite{fang2018nonlinear,feng2023review}, with the Extended Kalman filter (EKF) and the Unscented Kalman filter (UKF) being the most widely used in practice. The EKF handles nonlinearity by linearizing the transition and measurement functions about the current estimate through their Jacobians. This linearization entails several drawbacks \cite{wan2000unscented, julier2004unscented,rigatos2011derivative, sarkka2013bayesian}. It requires the model functions to be differentiable and their Jacobians to be available in closed form, which excludes non-differentiable, tabular, or black-box models. Because it retains only a first-order approximation of the nonlinearity, its accuracy and consistency deteriorate as the dynamics grow more strongly nonlinear or the estimation error becomes large, which in turn narrows the range of systems to which it can be applied reliably. The UKF was introduced to overcome these issues~\cite{julier2004unscented,wan2000unscented}. Rather than linearizing, the UKF propagates a deterministic set of sigma points through the nonlinear functions to recover the first two moments of the transformed distribution. This captures the mean and covariance to higher order than the EKF's first-order linearization. Because the filter evaluates the model functions directly rather than differentiating them, it requires no Jacobians, which in turn lets it handle non-differentiable models. As in the EKF, an explicit posterior covariance is available at every step.
 
Despite these advantages, the accuracy and statistical validity of any such filter hinge on two modeling assumptions that are seldom satisfied in practice. The first is that the process and measurement noise covariances are known. In most cases, they are fixed offline to constant values, even though the true noise statistics are time-varying and, under heavy-tailed disturbances, far from Gaussian \cite{banerjee2013underwater,jia2020novel,zhu2021robust,roth2013student}. The second is that the transition function used by the filter matches the true system dynamics. The transition function available to the filter can deviate from the true dynamics because of unmodeled effects~\cite{li2026enhanced,vilavalls2022robust}, numerical discretization of continuous-time models~\cite{kulikova2022continuous}, and parameter uncertainty and model mismatch~\cite{abdallah2022robust,chatzi2009unscented}. When either assumption is violated, the filter degrades in two distinct ways. Its state estimate loses accuracy, and its reported covariance may become miscalibrated. Depending on the direction and structure of the misspecification, the posterior may be either overconfident or conservative~\cite{ge2023novel, zhu2017conservative, govaers2013covariance}. Reliable covariance estimates are important because the reported uncertainty is used downstream for data-association gating, sensor-fusion weighting, and decision making under risk. These challenges have motivated robust filtering approaches for state estimation under difficult measurement and dynamical conditions, including non-Gaussian noise, target maneuvers, and intermittent observations~\cite{zhang2025novel}. 
 
Recently, hybrid model-based/data-driven filters that embed neural networks within the filtering recursion have attracted growing attention \cite{feng2023review,shlezinger2025artificial,shlezinger2023model,revach2022kalmannet,revach2022unsupervised,choi2023split,buchnik2023latent,buchnik2024gsp, jouaber2021nnakf,xu2024ekfnet,ghosh2024danse, ni2024adaptive}. A representative example is KalmanNet, which replaces the analytical Kalman gain with a recurrent neural network learned from data \cite{revach2022kalmannet}. By learning the gain directly, KalmanNet removes the dependence on explicit noise statistics and compensates for incomplete physics, achieving accurate tracking on partially known systems while retaining the low complexity of the underlying recursion. Numerous variants have since extended this idea to smoothing \cite{revach2023rtsnet}, high-dimensional measurements \cite{buchnik2023latent}, fast adaptation \cite{chen2025maml,ni2024adaptive}, and related settings \cite{choi2023split,buchnik2024gsp,revach2022unsupervised}. Recent studies have further explored model-based deep learning for dynamic state estimation and neural-aided Kalman filtering for maneuvering-target tracking, demonstrating growing interest in combining analytical state-space structure with learned adaptation \cite{lin2025attentional, mari2025ensemble}

These hybrid filters, however, share limitations that constrain their use as uncertainty-aware estimators. First, most are built on the EKF, which inherit its dependence on local linearization. Propagating the second-order moments requires Jacobians of the transition and measurement functions, which limits their applicability when these derivatives are unavailable, as in non-differentiable, tabular, or black-box models. Second, the original KalmanNet~\cite{revach2022kalmannet} learns the measurement update without maintaining an explicit posterior-covariance recursion and therefore provides a state estimate without an accompanying posterior covariance. A subsequent method reconstructs the error covariance from the learned gain and the known observation model, but requires the observation matrix, or the corresponding measurement Jacobian in nonlinear systems, to have full column rank~\cite{klein2022uncertainty}. This condition excludes common partially observed settings in which the number of measurements is smaller than the state dimension.

Recent work further improves estimation accuracy. EnKalmanNet \cite{li2026enhanced}, for instance, jointly learns the Kalman gain and a data-driven correction to the process model, reducing the prior error from both the prediction and update stages. Because it is trained with a state-estimation objective and does not maintain an explicit covariance recursion, however, it does not report or calibrate a posterior covariance. A complementary line of work restores explicit uncertainty information. Cholesky-KalmanNet~\cite{ko2024cholesky} and Eigen Decomposition-KalmanNet~\cite{ko2025robust} directly estimate the predicted state and innovation covariances within an EKF-based formulation, but retain its linearized structure and do not include a dedicated correction for gain bias induced by transition-model mismatch. Bayesian KalmanNet~\cite{dahan2025bayesian} instead estimates error covariance through stochastic sampling of the KalmanNet parameters, retaining KalmanNet's learned gain but not an explicit analytical covariance recursion. Consequently, existing methods do not jointly adapt the noise covariances, correct the analytical gain under model mismatch, and maintain an explicit nonlinear covariance recursion whose calibration can be directly optimized.
 
These observations motivate a different model-based deep learning strategy. Rather than replacing the analytical filtering structure, we integrate data-driven corrections into the UKF recursion,where nonlinear state and uncertainty information are propagated through sigma points and an explicit posterior covariance is retained. We propose \emph{Unscented KalmanNet} (UKN), a hybrid recursive estimator that combines the analytical UKF backbone with two structurally distinct learned components and a calibration-aware training objective. Neural networks have been combined with the UKF before, for example to learn the model correction~\cite{zhan2006nn} and the transition dynamics~\cite{yu2023unscented} or to adapt the noise statistics \cite{tobaly2025integrating,levy2026adaptive}. To the best of our knowledge, however, UKN is the first to embed KalmanNet-style discriminative gain learning within the sigma-point recursion while preserving an explicit, calibrated posterior covariance. Because it is built on the UKF rather than the EKF, UKN requires no Jacobians, which allows it to handle non-differentiable models. Just as important, it retains the explicit posterior covariance that makes calibration measurable and directly optimizable at every step. In this way, UKN retains model-based nonlinear uncertainty propagation while using data-driven adaptation to account for information that is not adequately represented by the nominal model and noise statistics.

UKN allocates the two adaptation roles to separate modules. NoiseNet, a recurrent module, replaces the user-specified process and measurement covariances with time-varying estimates predicted from the filter's own statistics. Its parameterization keeps both matrices symmetric positive definite at every step and reduces to the nominal filter at initialization. GainNet, also recurrent, applies a bounded residual correction to the analytical UKF gain, compensating the bias induced by transition-model mismatch without discarding the sigma-point moment structure on which the update is built.

The two modules are trained jointly under a composite objective that supplements the mean-squared state error with a calibration term on the posterior variances and a heavy-tailed term on the innovation. Weights on the two consistency terms adapt across training epochs according to filter-consistency diagnostics, so that accuracy and calibration are optimized together without manual balancing. We evaluate the estimator on four examples, three synthetic systems and one based on real drone-flight data, and show that it improves both state accuracy and posterior calibration over the analytical UKF and the existing neural network-aided Kalman filters.

The remainder of the paper is organized as follows. Section~\ref{sec:background} reviews the nonlinear state-space model, the UKF, and the limitations that motivate UKN. Section~\ref{sec:ukn} develops the two learned components and the training objective. Section~\ref{sec:experiments} reports the numerical examples, and Section~\ref{sec:conclusion} concludes.

\section{Background and Problem Setup}
\label{sec:background}

\subsection{System model and estimation objective}
\label{sec:model}

We consider the discrete-time nonlinear state-space model given by:
\begin{align}
    \bm{x}_{t} &= \bm{f}(\bm{x}_{t-1},\bm{u}_{t}) + \bm{w}_{t},
\label{eq:dyn}\\
    \bm{y}_{t} &= \bm{h}(\bm{x}_{t},\bm{u}_{t}) + \bm{v}_{t},
\label{eq:meas}
\end{align}
for $t = 1,2,\ldots,T$, where $\bm{x}_t \in \mathbb{R}^{n_x}$ is
the state, $\bm{u}_t \in \mathbb{R}^{n_u}$ is a known input, and
$\bm{y}_t \in \mathbb{R}^{n_y}$ is the measurement. The process function $\bm{f}$ and the measurement function $\bm{h}$ are nonlinear, and
are not required to be differentiable. The process disturbance $\bm{w}_t$ and the measurement disturbance $\bm{v}_t$ are mutually independent zero-mean random variables with covariances
\begin{equation}
    \mathbb{E}[\bm{w}_{t}\bm{w}_{t}^{\!\top}] = \bm{Q}_{t}, \qquad
    \mathbb{E}[\bm{v}_{t}\bm{v}_{t}^{\!\top}] = \bm{R}_{t},
\label{eq:noise}
\end{equation}
where $\bm{Q}_t,\bm{R}_t\succ\bm{0}$ are time-varying and unknown to the filter. No further distributional assumptions are imposed on $\bm{w}_t$ or $\bm{v}_t$.

The input $\bm{u}_t$ in Eq.~\eqref{eq:dyn} is sampled at time $t$ and drives the transition from $\bm{x}_{t-1}$ to $\bm{x}_t$. It is jointly observed with the measurement $\bm{y}_t$. This convention is consistent with end-of-step discretization of the underlying continuous-time dynamics, which is the standard treatment in many engineering applications.

The process function $\bm{f}$ in Eq.~\eqref{eq:dyn} is the nominal model available to the filter, and is, in general, not identical to the true process function $\bm{f}^{\star}$ that governs the data. The discrepancy can be written as:
\begin{equation}
    \bm{f}^{\star}(\bm{x},\bm{u}) \;=\; \bm{f}(\bm{x},\bm{u}) + \bm{\delta}(\bm{x},\bm{u}),
    \label{eq:mismatch}
\end{equation}
where $\bm{\delta}$ is an unknown residual that captures unmodeled nonlinearities, discretization error, inaccurate values of the nominal model parameters, and exogenous disturbances not represented by $\bm{u}_t$. The residual $\bm{\delta}$ is not random, but it is unknown to the filter. Throughout this work, the measurement function $\bm{h}$ is assumed to be correctly specified. Accordingly, incomplete physics refers to mismatch in the transition function $\bm{f}$, whereas departures in the measurement statistics are represented through $\bm{v}_t$ and the unknown covariance $\bm{R}_t$.

Given the measurements $\{\bm{y}_t\}_{t=1}^{T}$ and the inputs $\{\bm{u}_t\}_{t=1}^{T}$, the objective is to estimate the state $\bm{x}_t$ together with a measure of the uncertainty at each time $t$. In particular, the filter outputs a posterior mean $\hat{\bm{x}}_{t|t}$ and a posterior covariance $\hat{\bm{P}}_{t|t}$, required to satisfy two criteria: (i)~\emph{accuracy}, in the sense that the mean-squared error $\mathbb{E}\!\left[\|\bm{x}_t-\hat{\bm{x}}_{t|t}\|^{2}\right]$ is small; and (ii)~\emph{calibration}, in the sense that the posterior covariance is consistent with the actual estimation error, $\mathbb{E}\!\left[(\bm{x}_t-\hat{\bm{x}}_{t|t})(\bm{x}_t-\hat{\bm{x}}_{t|t})^{\!\top}\right]\approx \hat{\bm{P}}_{t|t}$. Posterior covariance calibration can be assessed using the normalized estimation error squared~(NEES), $\bm{e}_t^{\!\top} \hat{\bm{P}}_{t|t}^{-1} \bm{e}_t$, where $\bm{e}_t = \bm{x}_t-\hat{\bm{x}}_{t|t}$, while innovation consistency can be assessed using the normalized innovation squared~(NIS), $\bm{\nu}_t^{\!\top}\bm{S}_t^{-1}\bm{\nu}_t$, where $\bm{\nu}_t = \bm{y}_t-\hat{\bm{y}}_{t|t-1}$ is the innovation and $\bm{S}_t$ is its covariance under the filter. Under the Gaussian consistency assumptions, the NEES and NIS follow chi-squared distributions with $n_x$ and $n_y$ degrees of freedom, respectively, and hence have expected values $n_x$ and $n_y$~\cite{barshalom2001estimation}. NEES requires the true state and is therefore used for evaluation, whereas NIS can be computed without state ground truth. For ensemble-based evaluation over $N$ independent test trajectories, let the NEES of trajectory $k$ at time $t$ be:
\begin{equation}
    \varepsilon_t^{(k)}
    =
    \bm{e}_t^{(k)\top}
    \hat{\bm{P}}_{t|t}^{(k)-1}
    \bm{e}_t^{(k)}.
\end{equation}
We report the average NEES (ANEES) as:
\begin{equation}
    \overline{\varepsilon}_t
    =
    \frac{1}{N}
    \sum_{k=1}^{N}
    \varepsilon_t^{(k)}.
\label{eq:anees}
\end{equation}
Accordingly, because the trajectory-wise NEES values are independent, their sum follows a chi-squared distribution with $Nn_x$ degrees of freedom:
\begin{equation}
    N\overline{\varepsilon}_t
    =
    \sum_{k=1}^{N}\varepsilon_t^{(k)}
    \sim \chi^2_{Nn_x}.
\end{equation}
The pointwise $(1-\alpha)$ ANEES consistency interval is therefore:
\begin{equation}
    \frac{\chi^2_{Nn_x,\alpha/2}}{N}
    \leq
    \overline{\varepsilon}_t
    \leq
    \frac{\chi^2_{Nn_x,1-\alpha/2}}{N}.
\label{eq:anees_interval}
\end{equation}
ANEES values above the upper limit indicate overconfidence, whereas values below the lower limit indicate conservative covariance estimates.

Two main challenges arise from the unknown $\{\bm{Q}_t\}, \{\bm{R}_t\}$ and the unknown residual $\bm{\delta}$ in Eq.~\eqref{eq:mismatch}. The next subsection reviews the unscented Kalman filter, which addresses the nonlinearity in $\bm{f}$ and $\bm{h}$ through a sigma-point representation but leaves these two challenges unresolved.

\subsection{The Unscented Kalman Filter}
\label{sec:ukf}

The Unscented Kalman Filter (UKF)~\cite{julier2004unscented,wan2000unscented} is a recursive estimator for nonlinear state-space models that propagates the first two moments of the state without linearization. The key idea, known as the Unscented Transform, is to represent the current belief about the state by a small set of deterministic sample points called \emph{sigma points}, propagate each sigma point through the nonlinear function, and recover the mean and covariance of the transformed distribution as a weighted sum over the propagated points. This approach captures the first two moments of the transformed distribution to at least second order in the Taylor expansion of the nonlinearity, regardless of differentiability.

At each time step, the UKF maintains a Gaussian approximation of the state posterior in the form of a mean and covariance, and proceeds in two stages. In the \emph{prediction stage}, $2n_x + 1$ sigma points generated from the posterior $(\hat{\bm{x}}_{t-1|t-1}, \hat{\bm{P}}_{t-1|t-1})$ at previous time $t-1$ are propagated through the process function $\bm{f}$ to estimate the prior $(\hat{\bm{x}}_{t|t-1}, \hat{\bm{P}}_{t|t-1})$ on the state at time $t$. In the \emph{update stage}, a new set of sigma points generated from the prior is propagated through the measurement function $\bm{h}$ to produce the predicted measurement $\hat{\bm{y}}_{t|t-1}$, the innovation covariance $\bm{S}_t$, and the state-measurement cross-covariance $\bm{C}^{xy}_t$. The measurement $\bm{y}_t$ is then folded into the posterior through the innovation $\bm{\nu}_t = \bm{y}_t - \hat{\bm{y}}_{t|t-1}$ and the Kalman gain $\bm{K}^{\mathrm{UKF}}_t = \bm{C}^{xy}_t \bm{S}_t^{-1}$, yielding the updated posterior $(\hat{\bm{x}}_{t|t}, \hat{\bm{P}}_{t|t})$. The full recursion is summarized in Algorithm~\ref{alg:ukf}.

\begin{algorithm}[!b]
\caption{Unscented Kalman Filter (UKF)}
\label{alg:ukf}
\begin{algorithmic}[1]
    \item[] \hspace{-1.5em} \textbf{Require:} posterior $(\hat{\bm{x}}_{t-1|t-1}, \hat{\bm{P}}_{t-1|t-1})$, input $\bm{u}_t$, measurement $\bm{y}_t$, covariances $\bm{Q}_t$ and $\bm{R}_t$, scaling parameter $\lambda$

    \item[] \vspace{0.5em} \hspace{-1.5em} \textbf{Prediction:}
    \STATE \hspace{0.3cm} $\bm{\chi}_0 \gets \hat{\bm{x}}_{t-1|t-1}$
    \STATE \hspace{0.3cm} $\bm{\chi}_i \gets \hat{\bm{x}}_{t-1|t-1} \pm
      \bigl(\sqrt{(n_x+\lambda)\bm{P}_{t-1|t-1}}\bigr)_i$, \qquad  $i=1,\ldots,n_x$
    \STATE \hspace{0.3cm} $\bm{\chi}_i^{\star} \gets \bm{f}(\bm{\chi}_i, \bm{u}_t)$,\ \
      $i=0,\ldots,2n_x$
    \STATE \hspace{0.3cm} $\hat{\bm{x}}_{t|t-1} \gets \sum_{i=0}^{2n_x} W^{(m)}_i
      \bm{\chi}_i^{\star}$
    \STATE \hspace{0.3cm} $\hat{\bm{P}}_{t|t-1} \gets \sum_{i=0}^{2n_x} W^{(c)}_i
      (\bm{\chi}_i^{\star} - \hat{\bm{x}}_{t|t-1})
      (\bm{\chi}_i^{\star} - \hat{\bm{x}}_{t|t-1})^{\!\top} + \bm{Q}_t$

    \item[] \vspace{0.5em} \hspace{-1.5em} \textbf{Update:}
    \STATE \hspace{0.3cm} Regenerate $\{\bm{\chi}_i\}$ from
      $(\hat{\bm{x}}_{t|t-1}, \hat{\bm{P}}_{t|t-1})$
    \STATE \hspace{0.3cm} $\bm{\gamma}_i \gets \bm{h}(\bm{\chi}_i, \bm{u}_t)$,\ \
      $i=0,\ldots,2n_x$
    \STATE \hspace{0.3cm} $\hat{\bm{y}}_{t|t-1} \gets \sum_i W^{(m)}_i \bm{\gamma}_i$
    \STATE \hspace{0.3cm} $\bm{S}_t \gets \sum_i W^{(c)}_i
      (\bm{\gamma}_i - \hat{\bm{y}}_{t|t-1})
      (\bm{\gamma}_i - \hat{\bm{y}}_{t|t-1})^{\!\top} + \bm{R}_t$
    \STATE \hspace{0.3cm} $\bm{C}^{xy}_t \gets \sum_i W^{(c)}_i
      (\bm{\chi}_i - \hat{\bm{x}}_{t|t-1})
      (\bm{\gamma}_i - \hat{\bm{y}}_{t|t-1})^{\!\top}$
    \STATE \hspace{0.3cm} $\bm{K}^{\mathrm{UKF}}_t \gets \bm{C}^{xy}_t \bm{S}_t^{-1}$
    \STATE \hspace{0.3cm} $\bm{\nu}_t \gets \bm{y}_t - \hat{\bm{y}}_{t|t-1}$
    \STATE \hspace{0.3cm} $\hat{\bm{x}}_{t|t} \gets \hat{\bm{x}}_{t|t-1} +
      \bm{K}^{\mathrm{UKF}}_t \bm{\nu}_t$
    \STATE \hspace{0.3cm} $\hat{\bm{P}}_{t|t} \gets \hat{\bm{P}}_{t|t-1} -
      \bm{K}^{\mathrm{UKF}}_t \bm{S}_t
      (\bm{K}^{\mathrm{UKF}}_t)^{\!\top}$
    \STATE \hspace{0.3cm} \textbf{return}  $(\hat{\bm{x}}_{t|t}, \hat{\bm{P}}_{t|t})$
\end{algorithmic}
\end{algorithm}

The sigma points and their weights are determined by three tuning parameters $\alpha$, $\beta$, and $\kappa$, where $\alpha > 0$ controls the spread of the sigma points around the mean, $\beta$ encodes prior knowledge about the underlying distribution, and $\kappa$ is a secondary scaling factor commonly set to $\kappa = 0$. The composite scaling parameter $\lambda$ in Algorithm~\ref{alg:ukf} is then defined as $\lambda = \alpha^2(n_x + \kappa) - n_x$. In this work, we set $\alpha = 1.0$, $\beta = 0$, and $\kappa = 0$.

\subsection{Limitations of the UKF under Incomplete Physics and Unknown Noise}
\label{sec:ukf_limits}

The UKF inherits limitations from two features implicit in Algorithm~\ref{alg:ukf}: (1) the noise covariances $\bm{Q}_t$ and $\bm{R}_t$ enter as user-specified inputs, and (2) the Kalman gain is derived under the assumption $\bm{\delta} = \bm{0}$. Three consequences follow.

First, \textbf{(L1)} the covariances $\bm{Q}_t$ and $\bm{R}_t$ must be specified by the user. In practice, they are typically unknswon and fixed to constant values obtained from offline analysis or heuristic defaults. In nonstationary settings, this choice is a compromise. Small values yield tight but overconfident posteriors. Large values yield loose but conservative ones. Neither tracks the actual time-varying uncertainty in the data.

Second, \textbf{(L2)} the analytical gain $\bm{K}^{\mathrm{UKF}}_t = \bm{C}^{xy}_t \bm{S}_t^{-1}$ is biased under incomplete physics. When the residual $\bm{\delta}$ in \eqref{eq:mismatch} is nonzero, the innovation $\bm{\nu}_t$ has two components: a random component from $\bm{w}_t$ and $\bm{v}_t$, and a systematic component originating from $\bm{\delta}$. The analytical $\bm{C}^{xy}_t$ and $\bm{S}_t$, however, are computed under the assumption $\bm{\delta} = \bm{0}$, and they do not account for this systematic component. The resulting gain either over- or under-weights the innovation relative to the gain that would be optimal for the true system.

Third, \textbf{(L3)} under incomplete physics, the analytical recursion does not generally satisfy accuracy and covariance calibration criteria (i) and (ii). The posterior covariance is updated as $\hat{\bm{P}}_{t|t} = \hat{\bm{P}}_{t|t-1} - \bm{K}^{\mathrm{UKF}}_t \bm{S}_t (\bm{K}^{\mathrm{UKF}}_t)^{\!\top}$, regardless of whether the gain is appropriate for the true system. Under incomplete physics, the gain corrects the state mean using the full innovation $\bm{\nu}_t$, which contains a systematic contribution from $\bm{\delta}$. Because $\bm{P}_{t|t}$ contracts by the nominal amount while the actual estimation error remains increased by the mismatch, the posterior becomes inconsistent with the actual error, in violation of criterion (ii).

These three limitations share a common cause. The recursion in Algorithm~\ref{alg:ukf} is closed-form and offers no mechanism to adapt either the noise covariances or the gain in response to the data. They also share a common implication. Improving accuracy and calibration are distinct objectives that can be difficult to address reliably through a single learnable quantity. They therefore call for structurally distinct adaptation mechanisms.

\section{Unscented KalmanNet}
\label{sec:ukn}

\begin{figure}[!t]
    \centering
    \includegraphics[width=\linewidth]{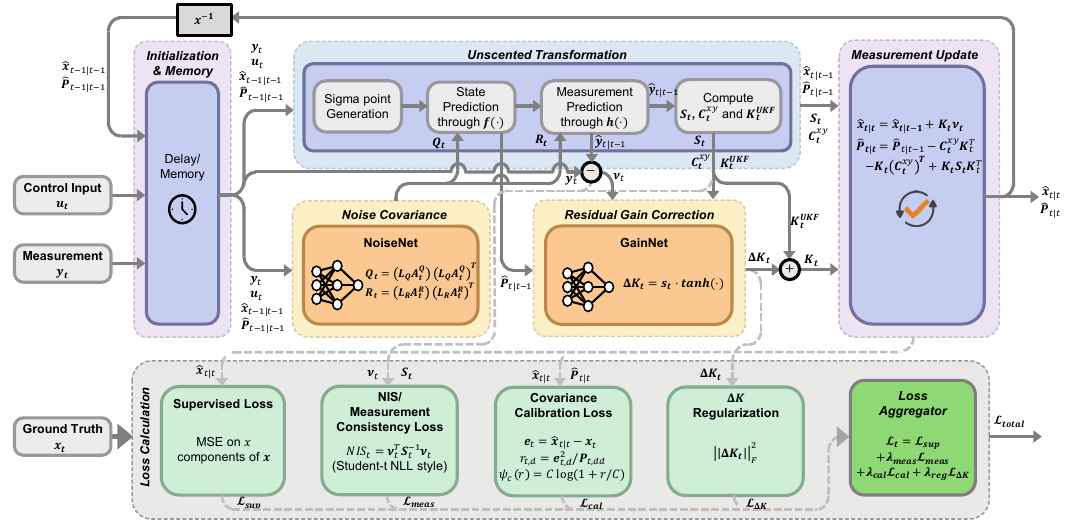}
    \vspace{-8pt}
    \caption{Architecture of the Unscented KalmanNet (UKN). A UKF backbone (\emph{Unscented Transform}) generates the sigma points and propagates them through the process and measurement functions $\bm{f}$ and $\bm{h}$ to produce the prior moments $\hat{\bm{x}}_{t|t-1}$, $\hat{\bm{P}}_{t|t-1}$, the predicted measurement $\hat{\bm{y}}_{t|t-1}$, the innovation covariance $\bm{S}_t$, the state-measurement cross-covariance $\bm{C}_t^{xy}$, and the analytical gain $K_t^{\mathrm{UKF}}$. Two learned modules adapt the recursion. \emph{NoiseNet} predicts the process and measurement covariances through Cholesky factors, $\bm{Q}_t= (\bm{L}_{Q,\mathrm{base}} \bm{A}^Q_t)(\bm{L}_{Q,\mathrm{base}} \bm{A}^Q_t)^{\!\top}$ and $\bm{R}_t = (\bm{L}_{R,\mathrm{base}} \bm{A}^R_t)(\bm{L}_{R,\mathrm{base}} \bm{A}^R_t)^{\!\top}$, which guarantees positive definiteness; \emph{GainNet} outputs a bounded residual $\Delta \bm{K}_t = s_t \tanh(\cdot)$ that is added to $\bm{K}_t^{\mathrm{UKF}}$ to form the corrected gain $\bm{K}_t$. The corrected gain is used in the posterior-mean update, while the posterior covariance is updated using the arbitrary-gain covariance identity. Equivalently, the nominal UKF posterior covariance is augmented by the positive-semidefinite term $\Delta\bm{K}_t\bm{S}_t\Delta\bm{K}_t^{\!\top}$. During training (bottom), a supervised MSE term $\mathcal{L}_{\mathrm{sup}}$, a Student-$t$ innovation-consistency term $\mathcal{L}_{\mathrm{meas}}$, a posterior-covariance calibration term $\mathcal{L}_{\mathrm{cal}}$, and a residual-gain regularizer $\mathcal{L}_{\Delta K}$ are combined by the loss aggregator with adaptive weights into   $\mathcal{L}_{\mathrm{total}}$.\vspace{-1.5em}}
  \label{fig:ukn_inference}
\end{figure}

\begin{figure*}[!t]
    \centering
    \includegraphics[width=0.8\linewidth]{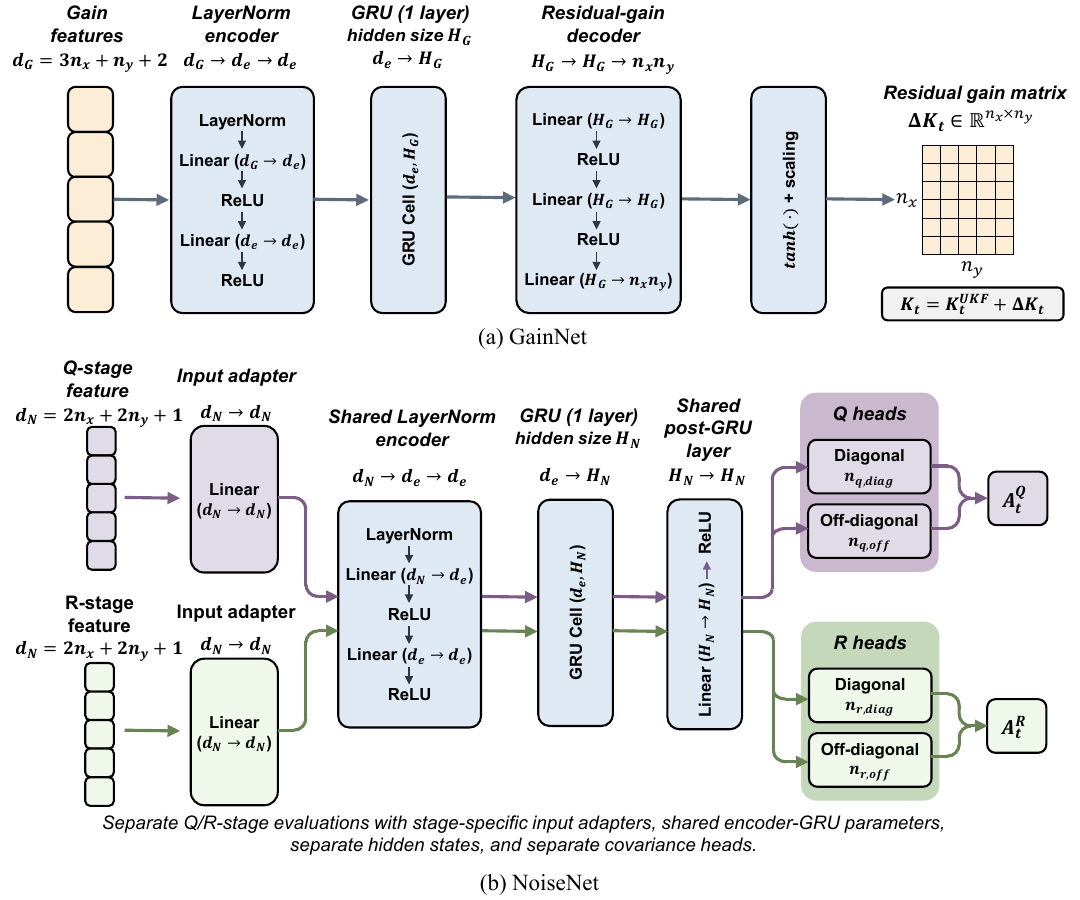}
    \caption{Architectures of GainNet and NoiseNet. (a) GainNet maps a $d_G$-dimensional vector of filter statistics through a layer-normalized encoder, a single-layer GRU, and a residual-gain decoder to produce $\Delta\bm K_t\in\mathbb{R}^{n_x\times n_y}$. (b) NoiseNet processes the Q- and R-stage features separately through stage-specific input adapters, while sharing the encoder-GRU parameters and maintaining separate recurrent hidden states and covariance heads. The input and output dimensions are determined by $n_x$ and $n_y$, whereas the encoder embedding dimension $d_e$ and the recurrent hidden dimensions $H_G$ and $H_N$ are specified for each numerical example.}
  \label{fig:network_architecture}
\end{figure*}

The Unscented KalmanNet (UKN) is a hybrid recursive estimator that combines two learned components and a calibration-aware training objective within the analytical UKF recursion of Algorithm~\ref{alg:ukf}. It is designed to address the three limitations identified above while preserving the sigma-point representation of the predicted and updated moments.

Fig.~\ref{fig:ukn_inference} illustrates the information flow within a single filtering step. The framework has three elements. First, a differentiable UKF backbone carries out the sigma-point prediction and update of Algorithm~\ref{alg:ukf}. Second, \emph{NoiseNet} is a recurrent module that produces the time-varying process and measurement covariances $\bm{Q}_t$ and $\bm{R}_t$ from the recent posterior and innovation. Third, \emph{GainNet} is a recurrent module that produces a bounded residual correction $\Delta\bm{K}_t$ to the analytical UKF gain $\bm{K}^{\mathrm{UKF}}_t$. At each time step, NoiseNet provides $\bm{Q}_t$ prior to the prediction stage and $\bm{R}_t$ prior to the update stage, while GainNet adjusts the analytical gain within the update. The resulting posterior and innovation feed NoiseNet at the next step.

The original KalmanNet~\cite{revach2022kalmannet} replaces the analytical Kalman gain with a single recurrent neural module, built on an EKF backbone that requires Jacobians of $\bm{h}$. Routing all corrections (e.g., noise misspecification, model mismatch, and covariance errors) through a single gain update creates a structural bottleneck, i.e., a single corrective quantity cannot independently address sources that affect accuracy and calibration in fundamentally different ways. In addition, the absence of an explicit covariance recursion leaves $\bm{P}_{t|t}$ undefined as a standalone output, so the filter provides a state estimate without a calibrated posterior covariance, leaving (L3) unaddressed. The proposed UKN resolves both issues by separating adaptation into \emph{NoiseNet} and \emph{GainNet} and retaining sigma-point moment propagation together with an explicit posterior-covariance recursion so that $\bm{P}_{t|t}$ remains an explicit quantity whose calibration can be evaluated and optimized.

The UKN allocates the adaptation responsibilities to two structurally distinct components. NoiseNet adapts the noise covariances $\bm{Q}_t$ and $\bm{R}_t$ and is trained with calibration-aware losses. GainNet then corrects the residual gain bias due to the model mismatch $\bm{\delta}$, which remains even when $\bm{Q}_t$ and $\bm{R}_t$ are correctly specified. This correction is bounded so that GainNet does not overwrite the analytical UKF gain. The analytical sigma-point moment propagation is retained as the backbone, with $\bm{f}$ and $\bm{h}$ continuing to act on the sigma points without modification. NoiseNet addresses (L1), GainNet addresses (L2), and the calibration-aware training objective binds them together to address (L3). All three components contribute to keeping accuracy and calibration in balance.

Fig.~\ref{fig:network_architecture} summarizes the internal architectures of the two learned modules. Both modules use a layer-normalized feature encoder followed by a single-layer GRU, but they differ in their inputs and output parameterizations. GainNet maps the current filter statistics to a residual gain matrix, whereas NoiseNet processes the Q- and R-stage features separately using stage-specific input adapters, shared encoder-GRU parameters, separate recurrent hidden states, and separate covariance heads.

\subsection{NoiseNet}
\label{sec:noisenet}
We begin with NoiseNet, which addresses limitation (L1) by replacing the user-specified covariances $\bm{Q}_t$ and $\bm{R}_t$ with time-varying estimates produced from the filter's own statistics (Fig.~\ref{fig:ukn_inference}). Earlier work on learning covariances within a filtering step has relied on direct regression \cite{ko2024cholesky, ko2025robust}. NoiseNet instead predicts bounded lower-triangular multipliers $\bm{A}^Q_t$ and $\bm{A}^R_t$ that act on the Cholesky factors of fixed baseline covariances:
\begin{equation}
    \begin{aligned}
        \bm{Q}_t &= (\bm{L}_{Q,\mathrm{base}} \bm{A}^Q_t)
                   (\bm{L}_{Q,\mathrm{base}} \bm{A}^Q_t)^{\!\top}, \\
        \bm{R}_t &= (\bm{L}_{R,\mathrm{base}} \bm{A}^R_t)
                   (\bm{L}_{R,\mathrm{base}} \bm{A}^R_t)^{\!\top},
    \end{aligned}
\label{eq:Q_R_chol}
\end{equation}
where $\bm{L}_{Q,\mathrm{base}}$ and $\bm{L}_{R,\mathrm{base}}$ are Cholesky factors of user-supplied baseline covariances $\bm{Q}_{\mathrm{base}}\succ \bm{0}$ and  $\bm{R}_{\mathrm{base}} \succ \bm{0}$. The product $\bm{L}_{Q,\mathrm{base}} \bm{A}^Q_t$ is itself a valid Cholesky factor whenever $\bm{A}^Q_t$ is lower triangular with positive diagonal. Eq.~\eqref{eq:Q_R_chol} therefore guarantees symmetric positive definite $\bm{Q}_t$ and $\bm{R}_t$ at every step and throughout training.

NoiseNet uses different features for predicting $\bm{Q}_t$ and $\bm{R}_t$, reflecting the different roles of the process and measurement covariances within a filtering step. The $Q$-stage runs before the prediction at time $t$ and uses information available from the posterior at time $t-1$:
\begin{equation}
    \bm{\phi}^Q_t = \Big[
    \bm{y}_{t-1}^{\!\top},\;
    \bm{\nu}_{t-1}^{\!\top},\;
    \hat{\bm{x}}_{t-1|t-1}^{\!\top},\;\log\mathrm{diag}(\hat{\bm{P}}_{t-1|t-1})^{\!\top},\;
    \|\bm{u}_{t-1}\|_2
    \Big]^{\!\top}.
\label{eq:phiQ}
\end{equation}
The $R$-stage runs before the update at time $t$ and uses the prior at time $t$:
\begin{equation}
    \bm{\phi}^R_t = \Big[
    \bm{y}_t^{\!\top},\;
    \bm{\nu}_t^{\!\top},\;
    \hat{\bm{x}}_{t|t-1}^{\!\top}, \; \log\mathrm{diag}(\hat{\bm{P}}_{t|t-1})^{\!\top},\;
    \|\bm{u}_t\|_2
    \Big]^{\!\top}.
\label{eq:phiR}
\end{equation}
The $Q$-stage feature asks how uncertain the upcoming transition should be, given the most recent posterior. The $R$-stage feature asks how reliable the current measurement is, given the propagated prior and the realized innovation. As shown in Fig.~\ref{fig:network_architecture}(b), the Q- and R-stage features are evaluated separately through stage-specific linear input adapters and a common layer-normalized encoder-GRU parameterization, while maintaining separate recurrent hidden states. The resulting features are passed to stage-specific output heads that emit the diagonal and strictly lower-triangular parameters of $\bm A_t^Q$ and $\bm A_t^R$.

The output of the network parameterizes the lower-triangular multipliers $\bm{A}^Q_t \in \mathbb{R}^{n_x \times n_x}$ and $\bm{A}^R_t \in \mathbb{R}^{n_y \times n_y}$. The diagonal entries are bounded by a log-domain saturation:
\begin{equation}
    [\bm{A}^Q_t]_{ii} =
    \sqrt{\mathrm{clip}\!\big(\exp(c_Q \tanh \tilde{q}_{t,i}),
    q_{\min}, q_{\max}\big)},
    \label{eq:Achol_diag}
\end{equation}
and the strictly lower-triangular entries are bounded by a direct tanh saturation,
\begin{equation}
    [\bm{A}^Q_t]_{ij} = c^{\mathrm{off}}_Q \tanh \tilde{a}^Q_{t,ij},
    \qquad i > j,
    \label{eq:Achol_off}
\end{equation}
with upper-triangular entries set to zero, where $\tilde{q}_{t,i}$ and $\tilde{a}^Q_{t,ij}$ are the decoder outputs for the diagonal and off-diagonal entries, respectively. Analogous expressions hold for $\bm{A}^R_t$ with constants $c_R$, $r_{\min}$, $r_{\max}$, and $c^{\mathrm{off}}_R$. The diagonal parameterization permits each direction of variance to scale between a fraction and a multiple of the baseline. In our implementation, the diagonal bounds $q_{\min} = r_{\min} = 0.1$ and $q_{\max} = r_{\max} = 50$ allow each variance direction to scale between one tenth and fifty times the baseline.

Three design choices in Eqs.~\eqref{eq:Q_R_chol}-\eqref{eq:Achol_off} are worth noting. First, the parameterization is multiplicative rather than additive. The learned correction acts as a relative deformation of the baseline Cholesky factor. The magnitude of the correction therefore adapts automatically to the scale of the baseline covariance. Second, the final layers of all covariance heads are initialized to zero, and the stage-specific input adapters are initialized to identity mappings. At initialization, $\bm{A}^Q_t = \bm{A}^R_t = \bm{I}$, so $\bm{Q}_t = \bm{Q}_{\mathrm{base}}$ and $\bm{R}_t = \bm{R}_{\mathrm{base}}$. NoiseNet therefore reproduces the nominal UKF at the start of training, and learns only relative covariance deformations supported by the data. Third, the diagonal saturation bounds in \eqref{eq:Achol_diag} prevent the predicted covariances from collapsing to zero or growing arbitrarily large. Such excursions would destabilize the inversion in the gain $\bm{S}_t^{-1}$ and propagate poorly conditioned moments through the recursion.

\subsection{GainNet}
\label{sec:gainnet}
GainNet addresses limitation (L2) by augmenting the analytical UKF gain with an elementwise-bounded residual correction $\Delta\bm{K}_t$ (Fig.~\ref{fig:ukn_inference}). Rather than replacing the analytical gain with an unconstrained neural estimator, UKN learns a residual correction around the analytical UKF gain:
\begin{equation}
\bm{K}_t
=
\bm{K}^{\mathrm{UKF}}_t
+
\Delta\bm{K}_t.
\label{eq:K_final}
\end{equation}
This residual formulation retains the sigma-point moments of the UKF as the reference measurement update while allowing a data-driven correction when the nominal gain is systematically suboptimal under transition-model mismatch.

Because the corrected gain does not, in general, satisfy $\bm{K}_t=\bm{C}^{xy}_t\bm{S}_t^{-1}$, the standard UKF covariance reduction $\hat{\bm{P}}_{t|t-1}-\bm{K}_t\bm{S}_t\bm{K}_t^\top$ cannot be applied directly. Motivated by the covariance identity for an arbitrary gain treated as non-random in the covariance expansion, we use the following plug-in second-moment update:
\begin{align}
    \hat{\bm{x}}_{t|t}
    &=
    \hat{\bm{x}}_{t|t-1}
    +
    \bm{K}_t\bm{\nu}_t,
    \label{eq:ukn_state_update}\\
    \hat{\bm{P}}_{t|t}
    &=
    \hat{\bm{P}}_{t|t-1}
    -
    \bm{C}^{xy}_t\bm{K}_t^\top
    -
    \bm{K}_t(\bm{C}^{xy}_t)^\top
    +
    \bm{K}_t\bm{S}_t\bm{K}_t^\top.
    \label{eq:ukn_cov_update}
\end{align}
In the UKN update, GainNet first produces the corrected gain from the current filter statistics, and the resulting matrix is then substituted into Eqs.~\eqref{eq:ukn_state_update} and \eqref{eq:ukn_cov_update} as a plug-in quantity. This calculation does not account explicitly for the statistical dependence of the corrected gain on the current innovation. The required second-order moments are approximated by the sigma-point quantities $\hat{\bm{P}}_{t|t-1}$, $\bm{S}_t$, and $\bm{C}^{xy}_t$. Eq.~\eqref{eq:ukn_cov_update} is therefore used as a plug-in second-moment approximation rather than as the exact covariance of the innovation-dependent estimator.

Substituting
$\bm{K}_t=\bm{K}^{\mathrm{UKF}}_t+\Delta\bm{K}_t$
and
$\bm{K}^{\mathrm{UKF}}_t=\bm{C}^{xy}_t\bm{S}_t^{-1}$
into Eq.~\eqref{eq:ukn_cov_update} gives the equivalent form
\begin{equation}
\hat{\bm{P}}_{t|t}
=
\hat{\bm{P}}^{\mathrm{UKF}}_{t|t}
+
\Delta\bm{K}_t\bm{S}_t\Delta\bm{K}_t^{\!\top},
\label{eq:ukn_cov_residual_form}
\end{equation}
where
\begin{equation}
\hat{\bm{P}}^{\mathrm{UKF}}_{t|t}
=
\hat{\bm{P}}_{t|t-1}
-
\bm{K}^{\mathrm{UKF}}_t
\bm{S}_t
(\bm{K}^{\mathrm{UKF}}_t)^{\!\top}.
\end{equation}
The additional term
$\Delta\bm{K}_t\bm{S}_t\Delta\bm{K}_t^{\!\top}$
is positive semidefinite. Therefore, the corrected covariance remains positive semidefinite whenever $\hat{\bm{P}}^{\mathrm{UKF}}_{t|t}$ is positive semidefinite and $\bm{S}_t$ is positive definite. The recursion reduces exactly to the nominal UKF when $\Delta\bm{K}_t=\bm{0}$.

GainNet takes as input a feature vector $\bm{\phi}^G_t$ that summarizes UKF-consistent filter statistics at time $t$:
\begin{equation}
    \bm{\phi}^G_t =
    \Big[
    \tilde{\bm{\nu}}_t^{\!\top},\;
    \mathrm{NIS}_t,\;
    \log\det(\bm{S}_t),\;
    \Delta\hat{\bm{x}}_{t|t-1}^{\!\top},\;
    \big(\log\mathrm{diag}(\hat{\bm{P}}_{t|t-1})\big)^{\!\top},\;
    (\bm{K}^{\mathrm{UKF}}_t\bm{\nu}_t)^{\!\top}
    \Big]^{\!\top}.
    \label{eq:gainfeat}
\end{equation}
where $\tilde{\bm{\nu}}_t = \bm{L}_{S,t}^{-1}\bm{\nu}_t$ is the whitened innovation, $\bm{L}_{S,t}$ is the Cholesky factor of $\bm{S}_t$, $\mathrm{NIS}_t = \tilde{\bm{\nu}}_t^{\!\top}\tilde{\bm{\nu}}_t$ is the normalized innovation squared, and $\Delta\hat{\bm{x}}_{t|t-1} = \hat{\bm{x}}_{t|t-1} - \hat{\bm{x}}_{t-1|t-1}$ is the predicted state increment. The features split into three groups. The first group ($\tilde{\bm{\nu}}_t$, $\mathrm{NIS}_t$, $\log\det\bm{S}_t$) quantifies the surprise of the current measurement. The second group ($\Delta\hat{\bm{x}}_{t|t-1}$, $\log\mathrm{diag}\hat{\bm{P}}_{t|t-1}$) characterizes the predicted state evolution and the scale of the prior uncertainty. The third group ($\bm{K}^{\mathrm{UKF}}_t \bm{\nu}_t$) is the nominal UKF state correction itself. As shown in Fig.~\ref{fig:network_architecture}(a), the feature vector is processed by a layer-normalized feature encoder, a single-layer GRU, and a residual-gain decoder. The decoder output is then mapped to an adaptively scaled, elementwise-bounded residual gain matrix,
\begin{equation}
\Delta\bm{K}_t = s_t \, \tanh\!\big(f_K(\bm{h}^G_t)\big),
\qquad
s_t = c_K \|\bm{K}^{\mathrm{UKF}}_t\|_F,
\label{eq:deltaK}
\end{equation}
where $\bm{h}^G_t$ is the GRU hidden state, $f_K(\cdot)$ is a
linear map onto $\mathbb{R}^{n_x \times n_y}$, $c_K > 0$ is a fixed scaling constant, and $\tanh(\cdot)$ is applied elementwise. Consequently, each entry of the residual gain satisfies
\begin{equation}
    \left|[\Delta\bm{K}_t]_{ij}\right| \le s_t = c_K\|\bm{K}^{\mathrm{UKF}}_t\|_F.
    \label{eq:deltaK_elementwise_bound}
\end{equation}
The factor $s_t$ provides an adaptive amplitude scale for the residual correction based on the magnitude of the analytical UKF gain. Together with the elementwise saturation, it prevents the learned correction from becoming excessively large while allowing its scale to vary with the nominal measurement update.

Two design choices in Eq.~\eqref{eq:deltaK} are worth noting. First, the scaling $s_t$ is proportional to the Frobenius norm of the analytical gain, rather than fixed. It is not intended as a hard Frobenius-norm constraint on the entire residual matrix. Instead, it adapts the saturation level of the GainNet output to the magnitude of the nominal update. When the analytical gain is small, the allowable residual entries are correspondingly small. When the analytical gain is large, proportionally larger residual entries are permitted while remaining limited by the adaptive scale $s_t$. Second, the final layer of $f_K$ is initialized to zero. At initialization, $\Delta\bm{K}_t = 0$ and $\bm{K}_t = \bm{K}^{\mathrm{UKF}}_t$. The UKN therefore behaves as the analytical UKF at the start of training, and the residual is learned only when supported by data. This initialization keeps the initial forward recursion identical to the analytical UKF and prevents random residual-gain outputs from destabilizing the early stages of training.

\subsection{Training Objective and Adaptive Weighting}
\label{sec:training}

The training objective addresses limitation (L3). Existing learned filters such as KalmanNet~\cite{revach2022kalmannet} are typically trained against mean squared error (MSE) on the state estimate $\hat{\bm{x}}_{t|t}$. MSE alone, however, says nothing about whether the estimated covariance $\hat{\bm{P}}_{t|t}$ is too small or too large, and can therefore lead to a miscalibrated posterior. We therefore introduce covariance-calibration and innovation-consistency losses, together with a residual-gain regularizer, and balance the consistency losses through weights that adapt during training.

The state-estimation MSE is given as:
\begin{equation}
    \mathcal{L}_{\mathrm{MSE}} =
    \mathbb{E}\!\left[\|\hat{\bm{x}}_{t|t} - \bm{x}_t\|_2^2\right].
    \label{eq:mse_loss}
\end{equation}
This term measures the accuracy of the state estimate against the ground truth. It does not, however, control the reported covariance $\hat{\bm{P}}_{t|t}$. Under MSE alone, $\hat{\bm{P}}_{t|t}$ can shrink toward zero without penalty.

The calibration loss aligns the marginal posterior variances, represented by the diagonal entries of the posterior covariance, with the empirical squared errors:
\begin{equation}
\mathcal{L}_{\mathrm{cal}} = \frac{1}{2}\,\mathbb{E}\!\left[\frac{1}{n_x}
\sum_{i=1}^{n_x}
\left\{
c\log\!\left(1 + \frac{e_{t,i}^2}{c\,[\hat{\bm{P}}_{t|t}]_{ii}}\right)
+ \log [\hat{\bm{P}}_{t|t}]_{ii}
\right\}\right],
\label{eq:cal_loss}
\end{equation}
where $e_{t,i} = \hat{x}_{t|t,i} - x_{t,i}$ and the constant $c > 0$ soft-clamps the ratio $e_{t,i}^2/[\hat{\bm{P}}_{t|t}]_{ii}$ for numerical stability. Because the loss is applied componentwise to the diagonal entries of $\hat{\bm{P}}_{t|t}$, it directly promotes calibration of the marginal posterior variances, rather than the full multivariate covariance structure. Eq.~\eqref{eq:cal_loss} reduces to the diagonal Gaussian negative log-likelihood when $e_{t,i}^2 \ll c[\hat{\bm{P}}_{t|t}]_{ii}$ and grows only logarithmically for larger normalized errors, thereby reducing the influence of extreme deviations. A multivariate extension could additionally incorporate terms involving $\log\det\hat{\bm{P}}_{t|t}$ and $\bm{e}_t^{\top}\hat{\bm{P}}_{t|t}^{-1}\bm{e}_t$ to account for the joint posterior covariance structure, but is not considered in the present study.

The measurement loss matches the realized innovation distribution to a multivariate Student-$t$ target with $\nu_{\mathrm{df}}$ degrees of freedom:
\begin{equation}
    \mathcal{L}_{\mathrm{meas}} = \frac{1}{2}\,\mathbb{E}\!\left[    \log\det\bm{S}_t +
    (\nu_{\mathrm{df}} + n_y)\log\!\left(1 + \frac{\mathrm{NIS}_t}{\nu_{\mathrm{df}}}\right)
    \right].
    \label{eq:meas_loss}
\end{equation}
The heavy-tailed Student-$t$ target makes $\mathcal{L}_{\mathrm{meas}}$ robust to outlier innovations that would otherwise dominate a Gaussian negative log-likelihood through the $\bm{S}_t^{-1}$ term. 

The gain regularizer penalizes the Frobenius norm of the residual gain $\Delta\bm{K}_t$:
\begin{equation}
    \mathcal{L}_{\Delta K} =
    \mathbb{E}\!\left[\|\Delta\bm{K}_t\|_F^2\right].
    \label{eq:dk_loss}
\end{equation}
This term shrinks the residual correction toward zero unless the remaining objective terms favor a nonzero correction. It keeps the corrected gain $\bm{K}_t$ close to the analytical UKF gain $\bm{K}^{\mathrm{UKF}}_t$ except where supported by the training data.

The adaptive weights $w_{\mathrm{cal}}^{(e)}$ and $w_{\mathrm{meas}}^{(e)}$ are driven by two filter consistency metrics. The calibration discrepancy $g_{\mathrm{cal}}^{(e)} = |\log(\overline{e^2} / \overline{P})|$ is the log-ratio of the mean squared error to the mean predicted variance, computed over the epoch. It vanishes when the epoch-averaged squared error matches the epoch-averaged predicted variance. The measurement consistency $g_{\mathrm{meas}}^{(e)} = |\log(\overline{\mathrm{NIS}}/n_y)|$ compares the empirical NIS mean to its theoretical value $n_y$ under correctly specified $\bm{S}_t$. Both metrics are smoothed across epochs by an exponential moving average,
\begin{equation}
    \bar{g}^{(e)} = \beta\,\bar{g}^{(e-1)} + (1-\beta)\,g^{(e)},
\label{eq:ema}
\end{equation}
where $\beta \in (0,1)$ controls the smoothing horizon. Each weight is then updated multiplicatively to drive $\bar{g}^{(e)}$ toward a target level $\tau$:
\begin{equation}
    w^{(e+1)} = \mathrm{clip}\!\left(
    w^{(e)}\exp [\eta(\bar{g}^{(e)} - \tau)],\,
    w_{\min},\,w_{\max}
    \right).
    \label{eq:w_update}
\end{equation}
The constant $\eta > 0$ controls the update rate. When the smoothed discrepancy exceeds the target $\tau$, the exponent is positive and $w$ increases, increasing the contribution of the corresponding loss term. When it falls below $\tau$, $w$ decreases.

The four loss terms are combined into a single total loss for
epoch $e$,
\begin{equation}
    \mathcal{L}_{\mathrm{total}}^{(e)} =
    \mathcal{L}_{\mathrm{MSE}}
    + r_{\mathrm{cal}}^{(e)} w_{\mathrm{cal}}^{(e)} \mathcal{L}_{\mathrm{cal}}
    + r_{\mathrm{meas}}^{(e)} w_{\mathrm{meas}}^{(e)} \mathcal{L}_{\mathrm{meas}}
    + w_{\Delta K} \mathcal{L}_{\Delta K},
\label{eq:total_loss}
\end{equation}
where $r_{\mathrm{cal}}^{(e)},\; r_{\mathrm{meas}}^{(e)} \in [0,1]$ are linear warmup ramps that gate the calibration and measurement losses during the early epochs. The MSE loss alone first establishes a reasonable mean estimate. The calibration and measurement losses are then introduced once the posterior provides meaningful $\hat{\bm{P}}_{t|t}$ and $\bm{S}_t$. After warmup, the adaptive weights in Eq.~\eqref{eq:w_update} balance the composite objective during training, reducing the need for manual loss balancing.

\section{Numerical Examples}
\label{sec:experiments}
We evaluate the UKN on four examples that progressively increase the difficulty of nonlinear state estimation and uncertainty quantification. The Lorenz attractor isolates transition-model mismatch by using correctly specified process and measurement noise covariances. The Duffing oscillator adds time-varying parameter mismatch, process-noise misspecification, and measurement outliers. The coordinated-turn tracking problem combines incomplete physics, underestimated process and measurement covariances, and heavy-tailed glint errors in a partially observed radar setting. Finally, the UZH-FPV drone dataset tests the method on real flight data, where the sources of model error and noise misspecification are unknown. Together, these examples assess whether the UKN can improve point-estimation accuracy while maintaining a reliable posterior covariance across controlled synthetic settings and real measurements.

Fig.~\ref{fig:network_architecture} summarizes the common architecture of GainNet and NoiseNet used across the numerical examples. The example-specific recurrent hidden dimensions $H_G$ and $H_N$ are summarized in Table~\ref{tab:network_dimensions}.

\begin{table}[!t]
\centering
\caption{Network dimensions used in the numerical examples.}
\label{tab:network_dimensions}
\begin{tabular}{lccccc}
\hline
Example & $(n_x,n_y)$ & $(d_G,d_N)$ & $d_e$ & $H_G$ & $H_N$ \\
\hline
Lorenz~(\ref{sec:lorenz_problem_setting}) & $(3,2)$ & $(13,11)$ & 32 & 32 & 32 \\
Duffing~(\ref{sec:exp_duffing}) & $(2,2)$ & $(10,9)$ & 32 & 32 & 32 \\
Maneuvering target~(\ref{sec:exp_ct}) & $(5,2)$ & $(19,15)$ & 32 & 128 & 64 \\
UZH-FPV~(\ref{sec:exp_uzh_fpv}) & $(6,3)$ & $(23,19)$ & 32 & 64 & 64 \\
\hline
\end{tabular}
\end{table}

\subsection{Lorenz Attractor with Partial Observations and Parameter Mismatch}
\label{sec:lorenz_problem_setting}
The first example considers nonlinear state estimation for the Lorenz attractor. The example is chosen so that the filter faces three challenges simultaneously: chaotic nonlinear dynamics, partial observability, and a structured model mismatch between the true transition model and the nominal transition model available to the filter.

\subsubsection{True dynamics}
The state $\bm{x}_t = [x_{1,t}, \; x_{2,t},\; x_{3,t}]^{\top} \in \mathbb{R}^3$ consists of the three latent Lorenz state variables. The true continuous-time dynamics are given by:
\begin{equation}
\begin{aligned}
    \dot{x}_{1,t} &= \sigma (x_{2,t} - x_{1,t}), \\
    \dot{x}_{2,t} &= x_{1,t}(\rho - x_{3,t}) - x_{2,t}, \\
    \dot{x}_{3,t} &= x_{1,t} x_{2,t} - \beta x_{3,t},
\end{aligned}
\label{eq:lorenz_continuous}
\end{equation}
where the true parameters are $\sigma = 10$, $\rho = 28$, and $\beta = 8/3$. The system is discretized at $\Delta t = 0.02$~s for $T = 500$ time steps. No external control input is used.
 
\subsubsection{Nominal model and mismatch}
The filter uses a nominal transition model with fixed parameters
$(\bar{\sigma},\; \bar{\rho},\; \bar{\beta})=(8,\; 25,\; 8/3)$ instead of the true parameters. The nominal transition function $\bm{f}(\bm{x}_{t})$ is obtained by applying a fourth-order Runge-Kutta method with five internal substeps, using the nominal parameters in place of the true ones. Therefore, the simulated data and the filtering model evolve according to different Lorenz systems. The resulting residual $\bm{\delta}(\bm{x})=\bm{f}^{\star}(\bm{x})-\bm{f}(\bm{x})$ is state-dependent and can grow rapidly because of the chaotic sensitivity of the Lorenz dynamics.

\begin{figure*}[!t]
    \centering

    \includegraphics[width=\linewidth]{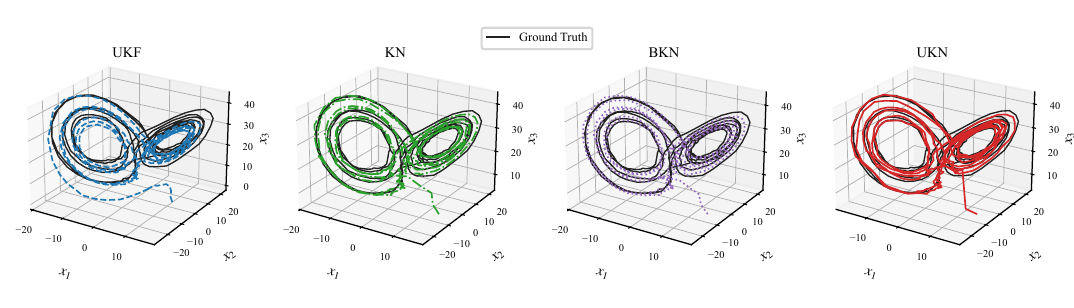}
    \vspace{-30pt}
    \caption{Estimated Lorenz trajectories in state space for a representative test sequence.}
    \vspace{-1.5em}
  \label{fig:lorenz1}
\end{figure*}

\begin{figure}[!t]
    \centering
    \includegraphics[width=0.55\linewidth]{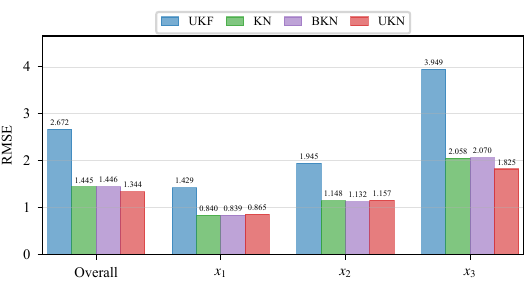}
    \vspace{-10pt}
    \caption{Per-component and overall RMSE on the test set.}
    \vspace{-1em}
  \label{fig:lorenz6}
\end{figure}

\begin{figure*}[!t]
    \centering
    \includegraphics[width=\linewidth]{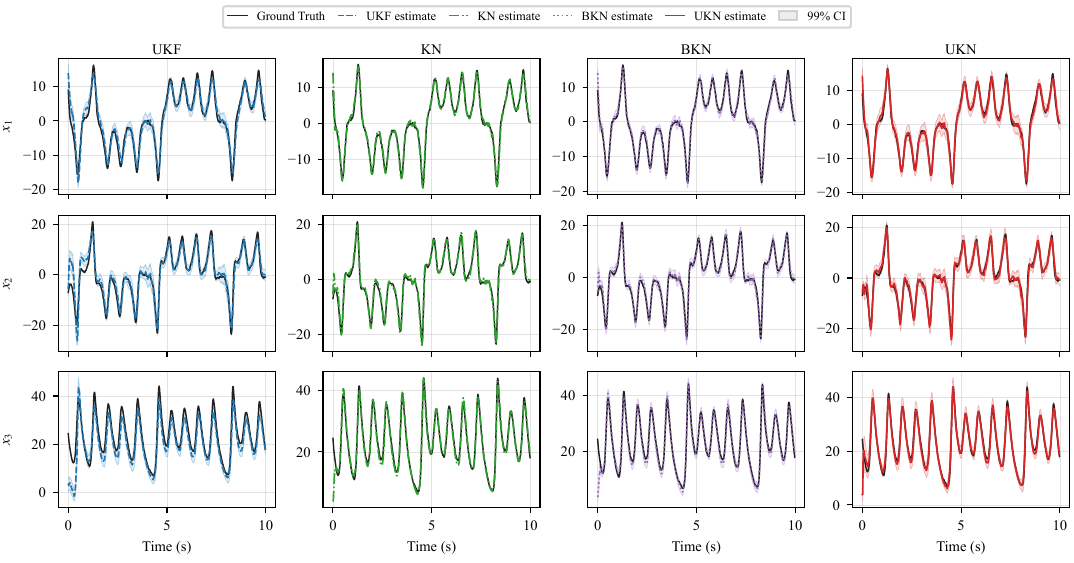}
    \vspace{-10pt}
    \caption{Estimated state components with marginal 99\% posterior intervals. The KN produces no posterior covariance and therefore no posterior interval.}
  \label{fig:lorenz2}
\end{figure*}

 \begin{figure*}[!t]
    \centering
    \includegraphics[width=\linewidth]{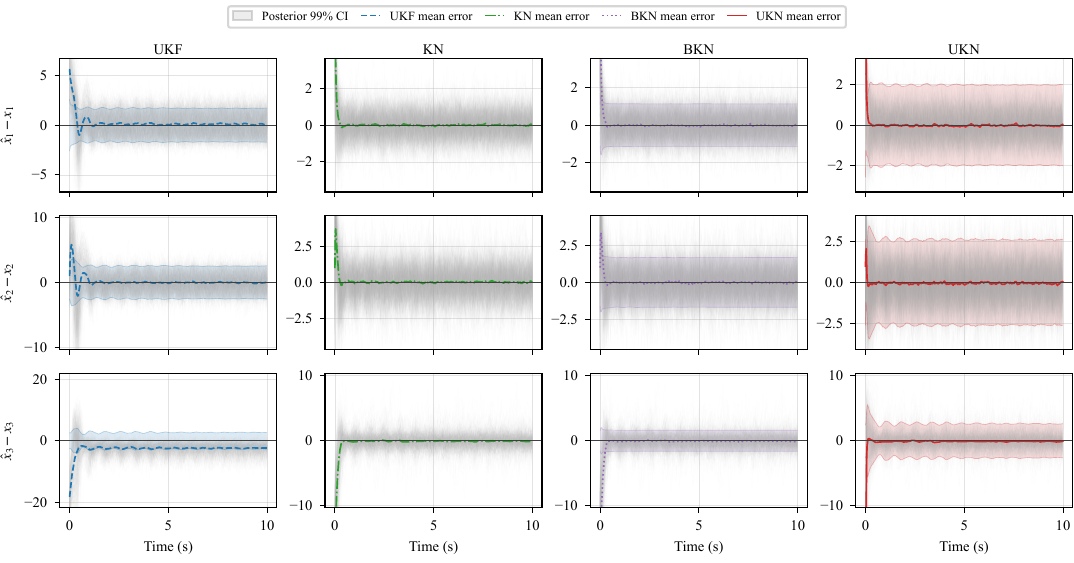}
    \vspace{-20pt}
    \caption{Per-trajectory estimation error $\hat{x}_{i,t}-x_{i,t}$, for the four filters. The gray traces show the realized error for every test trajectory. The KN produces no posterior covariance and therefore no confidence interval. \vspace{-2em}}
  \label{fig:lorenz3}
\end{figure*}

\subsubsection{Measurement model}
The measurement is two-dimensional and consists of a linear mixture of the three-dimensional latent state:
\begin{equation}
\bm{y}_t =
\begin{bmatrix}
1 & 0 & 0.3 \\
-0.2 & 1 & 0
\end{bmatrix}
\bm{x}_t + \bm{v}_t.
\label{eq:lorenz_meas}
\end{equation}
The measurement noise is zero-mean Gaussian:
\begin{equation}
\bm{v}_t \sim \mathcal{N}(\bm{0},\bm{R}^{\star}), \qquad
\bm{R}^{\star}=3.00^2\bm{I}_2.
\label{eq:lorenz_R}
\end{equation}
This observation model provides only two indirect measurements for the three-dimensional state. In particular, the third component $x_{3,t}$ contributes only weakly to the measurement (i.e., through the single coefficient $0.3$ in Eq.~\eqref{eq:lorenz_meas}). No outlier contamination is introduced.
 
\subsubsection{Process noise model}
After deterministic propagation over each sampling interval, additive zero-mean Gaussian process noise is applied to all state components:
\begin{equation}
\bm{w}_t \sim \mathcal{N}(\bm{0},\bm{Q}^{\star}), \qquad
\bm{Q}^{\star}=0.20^2\bm{I}_3.
\label{eq:lorenz_Q}
\end{equation}
The process noise introduces additional trajectory-to-trajectory variability on top of the sensitive deterministic evolution of the Lorenz system. As a result, the estimator must propagate uncertainty through a nonlinear chaotic model while also compensating for the systematic parameter mismatch in the nominal transition dynamics.
 
For this example we set $\bm{Q}_{\mathrm{base}} = \bm{Q}^{\star}$ and $\bm{R}_{\mathrm{base}} = \bm{R}^{\star}$, so that the noise covariances are correctly specified. The example therefore isolates robustness to transition-model mismatch in chaotic dynamics, with (L1) removed by construction.
 
\subsubsection{Initial condition and dataset}
The initial conditions are sampled independently for each trajectory according to:
\begin{equation}
x_{1,0} \sim \mathcal{U}[-10,10], \quad
x_{2,0} \sim \mathcal{U}[-10,10], \quad
x_{3,0} \sim \mathcal{U}[10,30].
\label{eq:lorenz_init}
\end{equation}
Independent training, validation, and test sets contain $2,400$, $300$, and $300$ trajectories, respectively. Consequently, this example evaluates whether the filter can reconstruct the full three-dimensional Lorenz state from two noisy and indirect observations while remaining robust to a systematic transition-model mismatch.

\begin{figure}[!t]
    \centering
    \includegraphics[width=0.55\linewidth]{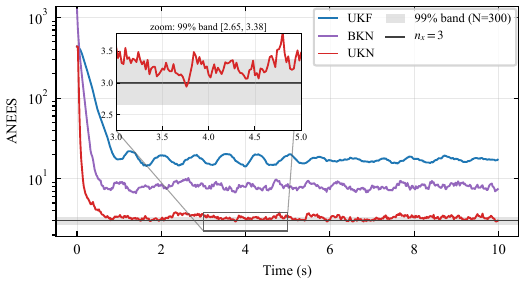}
    \vspace{-5pt}
    \caption{Average normalized estimation error squared (ANEES) over the $N=300$ test trajectories for the Lorenz example. The solid horizontal line denotes the nominal value $n_x=3$, and the horizontal shaded band denotes the pointwise 99\% ANEES consistency interval $[2.648,3.377]$.}
  \label{fig:lorenz5}
\end{figure}

\subsubsection{Results}
We benchmark the UKN against the analytical UKF~\cite{wan2000unscented} and two learned baselines, KalmanNet (KN)~\cite{revach2022kalmannet} and Bayesian KalmanNet (BKN)~\cite{dahan2025bayesian}. All four filters use the same nominal transition and measurement models; hence, the differences below reflect their respective update and covariance-modeling mechanisms rather than different physical assumptions.

Fig.~\ref{fig:lorenz1} shows the estimated trajectories in state space. The UKF deviates noticeably from the ground truth on the inner orbits and shows a prolonged transient deviation from the attractor manifold. The three learned filters reduce this deviation, with the UKN showing the shortest transient departure. This behavior indicates that the learned corrections mitigate a substantial portion of the systematic transition-model error induced by the parameter mismatch.

The aggregate RMSE in Fig.~\ref{fig:lorenz6} confirms this improvement. The overall RMSE decreases from $2.672$ for the UKF to $1.445$ for the KN, $1.446$ for the BKN, and $1.344$ for the UKN. The learned filters perform similarly on $x_1$ and $x_2$, which are the components most directly represented in the measurements. Their differences are concentrated in $x_3$, the most weakly observed component, where the RMSE decreases from $3.949$ for the UKF to $2.058$ for the KN, $2.070$ for the BKN, and $1.825$ for the UKN. Thus, the UKN's improvement over the learned baselines is concentrated primarily in $x_3$, the component least directly informed by the measurements.

The per-component trajectories and error distributions in Figs.~\ref{fig:lorenz2} and~\ref{fig:lorenz3} further show that the filters differ not only in accuracy but also in uncertainty representation. The UKF error in $x_3$ remains biased toward negative values, and its posterior interval does not consistently contain the true state. The BKN removes most of this bias but reports an interval that is narrower than the realized spread. In contrast, the UKN keeps the mean error close to zero while assigning an interval whose width is commensurate with the observed error. The KN achieves comparable point-estimate accuracy but provides no posterior covariance, so its uncertainty cannot be assessed.

As shown in Fig.~\ref{fig:lorenz5}, we evaluate covariance consistency using the ANEES in Eq.~\eqref{eq:anees}, computed over the $N=300$ independent test trajectories. Its expected value is $n_x=3$ in this example and its pointwise 99\% consistency interval from Eq.~\eqref{eq:anees_interval} is $[2.648,3.377]$. The UKF remains far above the upper limit, indicating significant overconfidence under the transition-model mismatch. The BKN also remains above the consistency region. In contrast, the UKN fluctuates around the nominal ANEES and remains within or close to the consistency region.

\subsection{Duffing Oscillator with Parameter Mismatch}
\label{sec:exp_duffing}
The second example considers a forced Duffing oscillator with trajectory-dependent stiffness parameters and an abrupt, unobserved change in the linear stiffness. In contrast to the previous example, where the noise covariances are correctly specified, this setting combines time-varying parameter mismatch with process-noise underestimation and outlier-contaminated nonlinear measurements. It therefore tests whether the filters can adapt to an unmodeled change in the dynamics while maintaining reliable uncertainty estimates under misspecified noise statistics.

\begin{figure*}[!t]
    \centering
    \includegraphics[width=\linewidth]{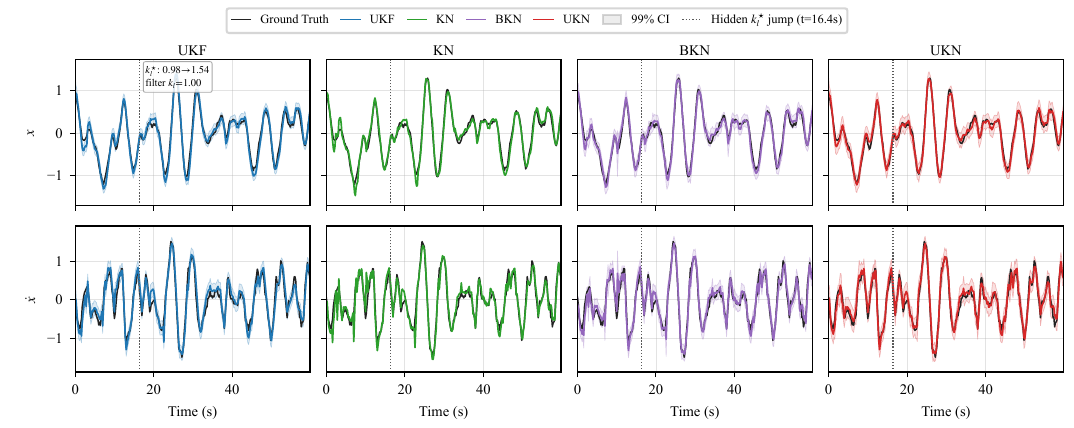}
    \vspace{-10pt}
    \caption{State estimation of displacement $x$ and velocity $\dot{x}$ for the Duffing oscillator with parameter mismatch on a representative test trajectory. The dotted line marks the hidden linear-stiffness jump at $t = 16.4$~s.}
    \vspace{-1.5em}
  \label{fig:duffing2}
\end{figure*}

\begin{figure}[!t]
    \centering
    \includegraphics[width=0.55\linewidth]{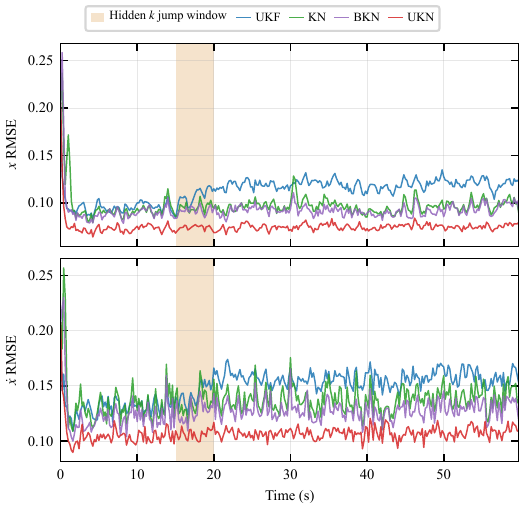}
    \vspace{-20pt}
    \caption{Time-resolved RMSE of each state component over the test set for the Duffing oscillator. \vspace{-2em}}
  \label{fig:duffing6}
\end{figure}

\begin{figure}[!t]
    \centering
    \includegraphics[width=0.55\linewidth]{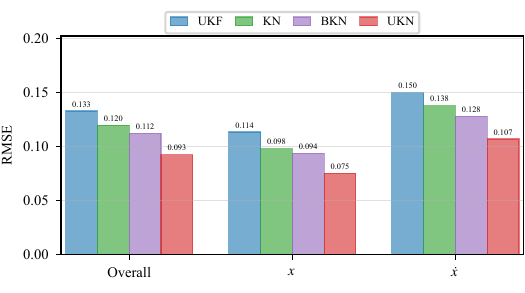}
    \vspace{-10pt}
    \caption{Per-component and overall RMSE on the test set for the Duffing oscillator.}
    \vspace{-1.5em}
  \label{fig:duffing4}
\end{figure}

\subsubsection{True dynamics}
The state $\bm{x}_t = [x_t, \dot{x}_t]^{\!\top} \in \mathbb{R}^2$ consists of the displacement $x_t$ and velocity $\dot{x}_t$. The true continuous-time dynamics are
\begin{equation}
    \ddot{x}_t + c\, \dot{x}_t
    + k_l^{\star} x_t
    + k_{nl}^{\star} x_t^3
    = u(t),
\label{eq:duffing_true}
\end{equation}
where $c=0.25$. For each trajectory, the stiffness parameters $(k_l^{\star}, k_{nl}^{\star})$ are sampled independently from $\mathcal{U}[0.8,1.2]\times\mathcal{U}[0.2,0.6]$. The nonlinear stiffness $k_{nl}^{\star}$ remains fixed throughout the trajectory, whereas the linear stiffness $k_l^{\star}$ undergoes an abrupt increase of 40-70\% at a time sampled from $\mathcal{U}[15,20]$~s. The system is integrated with time step $\Delta t=0.2$~s for $T=300$ steps. The external input $u(t)$ is piecewise constant over intervals of length $5\Delta t$, with each level drawn from $\mathcal{U}[-1,1]$.
 
\subsubsection{Nominal model and mismatch}
All filters use the same nominal transition model with fixed stiffness parameters $(\bar{k}_l,\bar{k}_{nl})=(1.0,0.4)$, corresponding to the means of the parameter distributions. The nominal transition $\bm{f}(\bm{x}_{t-1},\bm{u}_t)$ is obtained by Euler integration of Eq.~\eqref{eq:duffing_true}, with $(k_l^{\star},k_{nl}^{\star})$ replaced by $(\bar{k}_l,\bar{k}_{nl})$. Thus, the filters neither know the trajectory-specific stiffness values nor the subsequent jump in $k_l^{\star}$. The residual $\bm{\delta}(\bm{x},\bm{u})=\bm{f}^{\star}(\bm{x},\bm{u})- \bm{f}(\bm{x},\bm{u})$ is therefore state- and input-dependent, and its magnitude changes across trajectories and after the hidden stiffness jump.
 
\subsubsection{Measurement model}
The measurement is two-dimensional and nonlinear:
\begin{equation}
    \bm{y}_t =
        \begin{bmatrix}
            x_t + 0.1x_t^3 \\
            \dot{x}_t + 0.1\dot{x}_t^3
        \end{bmatrix}
    + \bm{v}_t .
\label{eq:duffing_meas}
\end{equation}
The measurement noise follows a contaminated Gaussian mixture, independently sampled at each step:
\begin{equation}
    \bm{v}_t \sim (1-p_{\mathrm{out}})
    \mathcal{N}(\bm{0},\sigma^2\bm{I}_2)
    +
    p_{\mathrm{out}}
    \mathcal{N}(\bm{0},\kappa\sigma^2\bm{I}_2),
    \label{eq:duffing_outlier}
\end{equation}
where $\sigma=0.15$, $p_{\mathrm{out}}=0.05$, and $\kappa=25$. The high-variance component represents intermittent outliers in both measurement channels.
 
\subsubsection{Noise covariance misspecification}
Additive zero-mean Gaussian process noise is applied at each step,
\begin{equation}
    \bm{w}_t \sim \mathcal{N}(\bm{0},\bm{Q}^{\star}), \qquad
    \bm{Q}^{\star}=\operatorname{diag}(10^{-3},10^{-2}),
\end{equation}
where the two entries correspond to displacement and velocity, respectively. The filters are instead given the baseline covariance
\begin{equation}
    \bm{Q}_{\mathrm{base}}=\operatorname{diag}(10^{-4},10^{-3}),
\end{equation}
which underestimates the true process variance by one order of magnitude in both components. The baseline measurement covariance is $\bm{R}_{\mathrm{base}}=\sigma^2\bm{I}_2$, matching only the nominal non-outlier component of Eq.~\eqref{eq:duffing_outlier}. It is therefore misspecified whenever an outlier occurs.
 
\subsubsection{Initial condition and dataset}
Each trajectory is initialized from
\begin{equation}
    x_0 \sim \mathcal{N}(1.0,0.04), \qquad
    \dot{x}_0 \sim \mathcal{N}(0,0.04).
\end{equation}
The filters are initialized at $\hat{\bm{x}}_{0|0}$ by approximately inverting Eq.~\eqref{eq:duffing_meas} at $\bm{y}_0$, with $\hat{\bm{P}}_{0|0}=\operatorname{diag}(0.04,0.04)$. Independent training, validation, and test sets contain $2400$, $300$, and $300$ trajectories, respectively.

\begin{figure*}[!t]
    \centering
    \includegraphics[width=\linewidth]{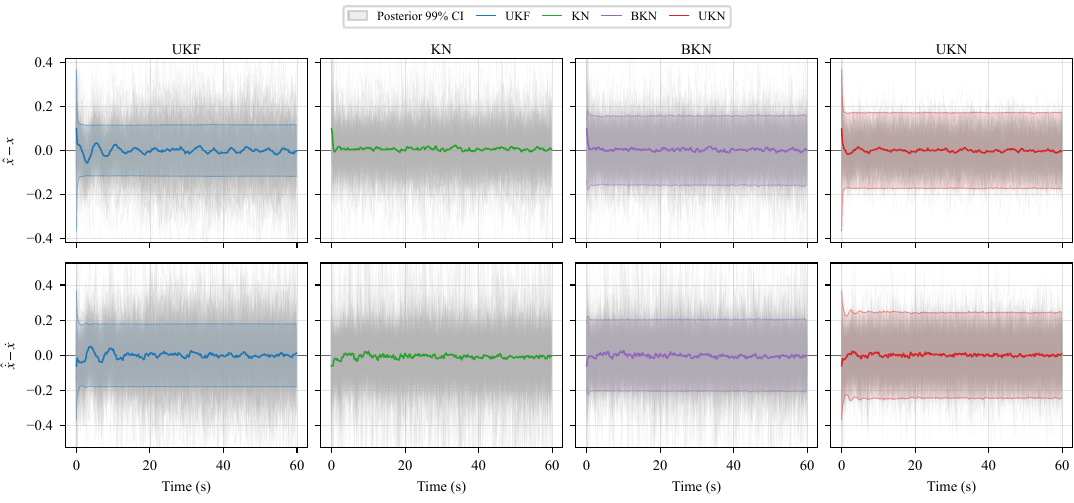}
    \vspace{-10pt}
    \caption{Per-component estimation error for the Duffing oscillator over the test set. The gray traces show the realized error for every test trajectory. The KN produces no posterior covariance and therefore no interval.}
    \vspace{-1em}
  \label{fig:duffing3}
\end{figure*}
 
\begin{figure}[!t]
    \centering
    \includegraphics[width=0.55\linewidth]{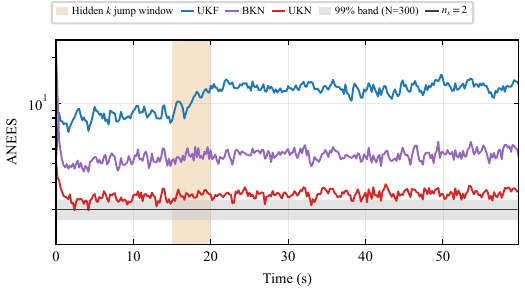}
    \vspace{-5pt}
    \caption{ANEES over the $N=300$ test trajectories for the Duffing oscillator. The solid horizontal line denotes the nominal value $n_x=2$, and the horizontal shaded band denotes the pointwise 99\% ANEES consistency interval $[1.715,2.310]$. The vertical shaded region denotes the interval over which the hidden linear-stiffness jump may occur.}
  \label{fig:duffing5}
\end{figure}
 
\subsubsection{Results}
Fig.~\ref{fig:duffing2} shows the state estimates on a representative test trajectory, where the hidden jump occurs at $t=16.4$~s and increases the linear stiffness from $0.98$ to $1.54$. All filters use the fixed nominal value $\bar{k}_l=1.00$ throughout and are not given either the jump time or its magnitude. Although all four filters track the displacement and velocity across the jump, their differences are difficult to assess from a single trajectory.

The time-resolved RMSE in Fig.~\ref{fig:duffing6} separates the filters more clearly. Before the jump window, the UKF, KN, and BKN have comparable errors, whereas the UKN is already the most accurate in both components. After the jump, the UKF error increases and remains elevated, reflecting the inability of the fixed nominal model to compensate for the changed stiffness. The learned filters are less affected. The KN and BKN largely preserve their displacement accuracy, while they show a moderate increase in velocity error. The UKN remains the most accurate filter throughout the sequence, indicating that its advantage is not restricted to the post-jump regime.

The aggregate RMSE in Fig.~\ref{fig:duffing4} shows the same ordering. The overall RMSEs are 0.133, 0.119, 0.112, and 0.093 for the UKF,
KN, BKN, and UKN, respectively. The UKN achieves a 30.1\% reduction relative to the UKF, which is substantially larger than the reductions achieved by the KN and BKN. Per component, the RMSE decreases from 0.114 to 0.075 (34.2\%) for $x$ and from 0.150 to 0.107 (28.7\%) for $\dot{x}$ when comparing the UKF with the UKN.

Fig.~\ref{fig:duffing3} compares the realized componentwise errors with the marginal 99\% posterior intervals obtained from the estimated covariance. The mean error of each filter remains close to zero in both components, indicating that the hidden stiffness change does not introduce a pronounced persistent bias. Among the covariance-reporting filters (i.e., UKF, BKN, and UKN), the UKN intervals most closely follow the empirical componentwise error spread. Again, the KN is excluded from the interval comparison because it does not provide a posterior covariance. These marginal intervals characterize each state component separately and do not fully determine joint covariance calibration, which also depends on the cross-covariance structure. The full-state covariance is therefore evaluated separately using ANEES.

Using the same ANEES criterion as in the previous example, Fig.~\ref{fig:duffing5} evaluates the full-state covariance over the $N=300$ test trajectories. Since $n_x=2$, the nominal ANEES is two, and the pointwise 99\% consistency interval is $[1.715,2.310]$. The UKF remains far above the upper limit and increases further following the hidden stiffness-change window, indicating severe underestimation of the joint estimation uncertainty. The BKN substantially reduces the ANEES relative to the UKF but remains above the consistency region throughout most of the sequence. The UKN produces the ANEES closest to the nominal value and substantially reduces the covariance miscalibration. Nevertheless, it remains moderately above the upper limit over much of the interval, indicating residual joint overconfidence rather than complete statistical consistency.


\subsection{Maneuvering Target Tracking under Model Mismatch}
\label{sec:exp_ct}

The third example considers a two-dimensional maneuvering target observed by a single range-and-bearing radar, a standard benchmark in nonlinear filtering~\cite{li2003survey,gustafsson2002particle}. The problem combines partial observability, unmodeled maneuver commands, underestimated process and measurement covariances, and heavy-tailed glint errors. Since only position-dependent quantities are measured, the speed, heading, and turn rate must be inferred indirectly from the accumulated motion. This example therefore provides a combined test of accuracy and covariance calibration under coupled model mismatch and non-Gaussian measurement errors.

\subsubsection{True dynamics}
The state
$\bm{x}_t=[p_{x,t},\,p_{y,t},\,v_t,\,\psi_t,\,\omega_t]^{\!\top} \in\mathbb{R}^5$
collects the planar position, speed, heading, and turn rate. The state evolves according to the coordinated-turn model~\cite{li2003survey}:
\begin{equation}
    \bm{x}_{t+1}=\bm{f}^{\star}(\bm{x}_t,\omega_t^{\mathrm{cmd}})+\bm{w}_t,
\end{equation}
where
\begin{equation}
    \bm{f}^{\star}(\bm{x}_t,\omega_t^{\mathrm{cmd}})=
    \begin{bmatrix}
        p_{x,t} + \tfrac{v_t}{\omega_t}\bigl[\sin(\psi_t + \omega_t \Delta t) - \sin\psi_t\bigr] \\[4pt]
        p_{y,t} + \tfrac{v_t}{\omega_t}\bigl[\cos\psi_t - \cos(\psi_t + \omega_t \Delta t)\bigr] \\[4pt]
        v_t \\[4pt]
        \psi_t + \omega_t \Delta t \\[4pt]
        (1-\rho)\omega_t + \rho\,\omega_t^{\mathrm{cmd}}
    \end{bmatrix}.
\label{eq:ctrv}
\end{equation}
The parameters are set to $\Delta t=1$~s and $\rho=0.15$. The constant-velocity limit is used whenever $|\omega_t|<10^{-4}$~rad/s to avoid the singularity in the position update.

The command $\omega_t^{\mathrm{cmd}}$ is piecewise constant over random segments and represents near-straight motion or left/right turns. It is used to generate the trajectories but is not provided to the filters. The process noise is Gaussian,
\begin{equation}
    \bm{w}_t \sim \mathcal{N}(\bm{0},\bm{Q}^{\star})
    \label{eq:ct_qtrue}
\end{equation}
where $\bm{Q}^{\star} =\operatorname{diag}\bigl(1.0^2,\,1.0^2,\,0.5^2,\,0.006^2,\,0.006^2\bigr)$.

\begin{figure}[!t]
    \centering
    \includegraphics[width=0.55\linewidth]{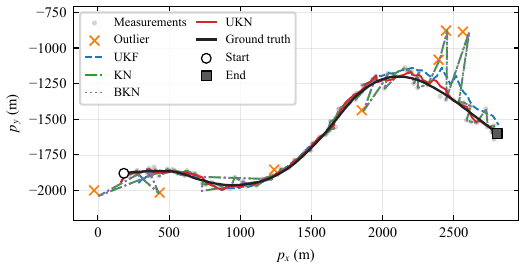}
    \vspace{-10pt}
    \caption{Maneuvering target tracking in the position plane for a representative test trajectory.}
    \vspace{-1em}
  \label{fig:ct1}
\end{figure}

\begin{figure*}[!t]
    \centering
    \includegraphics[width=\linewidth]{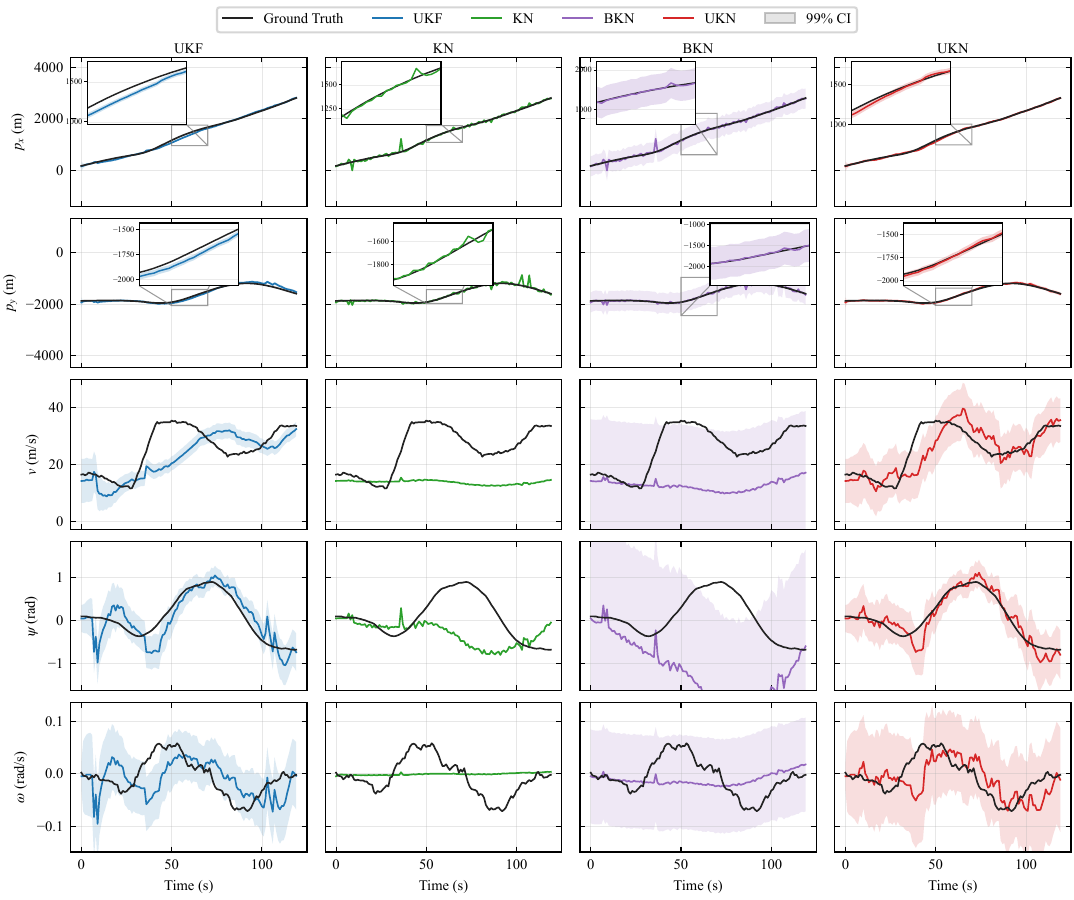}
    \vspace{-10pt}
    \caption{Estimated state components for the representative test trajectory of Fig.~\ref{fig:ct1} for the maneuvering target tracking. The shaded regions denote the marginal 99\% posterior intervals for the covariance-reporting filters.}
  \label{fig:ct2}
\end{figure*}

\subsubsection{Nominal model and mismatch}
All filters use the same nominal coordinated-turn model, but the hidden command $\omega_t^{\mathrm{cmd}}$ is unavailable. The nominal transition therefore replaces the last row of Eq.~\eqref{eq:ctrv} by
\[
\omega_{t+1}=\omega_t,
\]
so that the turn rate is treated as a random walk. The nominal and true transitions agree on the first four state components and differ only in the turn-rate component. The resulting deterministic residual is
\begin{equation}
\bm{\delta}(\bm{x}_t,t)
=
\bigl[
0,\;0,\;0,\;0,\;
\rho(\omega_t^{\mathrm{cmd}}-\omega_t)
\bigr]^{\!\top}.
\label{eq:ctrv_mismatch}
\end{equation}
This residual varies with the unobserved maneuver command and acts on the state component least directly constrained by the radar measurements.

\subsubsection{Measurement model}
A single radar at the origin provides range and bearing:
\begin{equation}
    \bm{y}_t=\bm{h}(\bm{x}_t)+\bm{v}_t,
    \qquad
    \bm{h}(\bm{x}_t)=
    \begin{bmatrix}
        r_t \\[2pt]
        \theta_t 
    \end{bmatrix}.
    \label{eq:ctrv_meas}
\end{equation}
where $r_t=\sqrt{p_{x,t}^2+p_{y,t}^2}$ and $\theta_t= \operatorname{atan2}(p_{y,t},p_{x,t})$. Thus, only position-dependent quantities are measured, while the speed, heading, and turn rate must be inferred indirectly through the dynamics.

In nominal operation, the radar noise is Gaussian,
\begin{equation}
    \bm{v}^{\mathrm{base}}_t \sim \mathcal{N}(\bm{0},\bm{R}^{\star}),\quad
    \bm{R}^{\star}=\operatorname{diag}\bigl((15~\mathrm{m})^2,\,(\pi/180)^2\bigr).
\end{equation}
To emulate glint, an additional impulsive error is added with probability
$p_{\mathrm{g}}=0.10$:
\begin{equation}
\bm{v}_t
=
\bm{v}^{\mathrm{base}}_t
+
g_t(\bm{b}_t+\bm{n}_t),
\qquad
g_t\sim\mathrm{Bernoulli}(p_{\mathrm{g}}).
\label{eq:ctrv_glint}
\end{equation}
The Bernoulli variable $g_t$ indicates whether a glint event occurs. When $g_t=0$, the measurement noise is the nominal Gaussian noise $\bm{v}^{\mathrm{base}}_t$; when $g_t=1$, the additional glint error $\bm{b}_t+\bm{n}_t$ is applied. The bias vector is $\bm{b}_t=[b_{r,t}, b_{\theta,t}]^{\top}$ where $b_{r,t}$ and $b_{\theta,t}$ denote the range and bearing biases, respectively. Their signs are sampled randomly, with $|b_{r,t}|\in[70\mathrm{m},160\mathrm{m}]$ and
$|b_{\theta,t}|\in[3\pi/180,8\pi/180]$. The additional glint scatter is
\begin{equation}
    \bm{n}_t\sim\mathcal{N}(\bm{0},\bm{\Sigma}_{\mathrm{g}}),\quad
    \bm{\Sigma}_{\mathrm{g}}=\operatorname{diag}\bigl((60~\mathrm{m})^2,\,(3\pi/180)^2\bigr).
\end{equation}
The resulting measurement noise is therefore heavy-tailed and nonstationary.

\subsubsection{Noise covariance misspecification}
The filters are not given the true process covariance in Eq.~\eqref{eq:ct_qtrue}. Instead, they use the nominal covariance 
\begin{equation}
    \bm{Q}^{\mathrm{nom}} = \operatorname{diag}\bigl(0.500,\,0.500, \,  6.13\times10^{-2},\,1.25\times10^{-5},\,1.80\times10^{-5}\bigr),
    \label{eq:ct_qnom}
\end{equation}
whose diagonal entries are smaller than the corresponding entries of $\bm{Q}^{\star}$. The nominal measurement covariance is
\begin{equation}
    \bm{R}^{\mathrm{nom}}=\operatorname{diag}\bigl((15\mathrm{m})^2,\,(\pi/180)^2\bigr),
\end{equation}
which matches the nominal Gaussian component of the radar noise but not
the glint-contaminated returns.

\subsubsection{Initial condition and dataset}
Each trajectory spans $T=120$ steps. The initial range and bearing are sampled as:
\begin{equation}
    r_0\sim\mathcal{U}[1500,3000]~\mathrm{m}, \qquad \phi_0\sim\mathcal{U}[-\pi,\pi].
\end{equation}
The initial speed and heading are sampled as:
\begin{equation}
    v_0\sim\mathcal{U}[15,35]~\mathrm{m/s}, \qquad \psi_0\sim\mathcal{U}[-\pi,\pi].
\end{equation}
The initial turn rate is sampled from a two-component mixture,
\begin{equation}
    \omega_0 \sim
    \begin{cases}
    \mathcal{U}[-0.005,0.005], & \text{with probability } 0.6,\\
    \mathcal{U}[0.015,0.06], & \text{with probability } 0.4,
    \end{cases}
\end{equation}
so that the initial conditions include both near-straight trajectories and moderate turning trajectories. All filters receive the same prior:
\begin{equation}
    \hat{\bm{x}}_0=\bm{x}_0+\bm{\varepsilon}_0,\qquad\bm{\varepsilon}_0\sim\mathcal{N}(\bm{0},\bm{P}_0),
\end{equation}
where $\bm{P}_0 =\operatorname{diag}\bigl(75^2,\, 75^2,\, 3^2,\, (7\pi/180)^2,\, 0.010^2 \bigr).$
Independent training, validation, and test sets contain $2400$, $300$, and $300$ trajectories, respectively.

\begin{figure}[!t]
    \centering
    \includegraphics[width=0.55\linewidth]{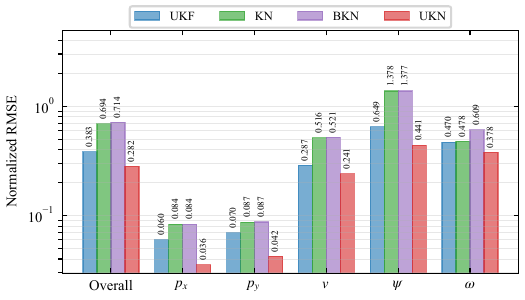}
    \vspace{-10pt}
    \caption{Per-component and overall normalized RMSE on the test set for the maneuvering target tracking.}
    \vspace{-1em}
  \label{fig:ct4}
\end{figure}

\begin{figure*}[!t]
    \centering
    \includegraphics[width=\linewidth]{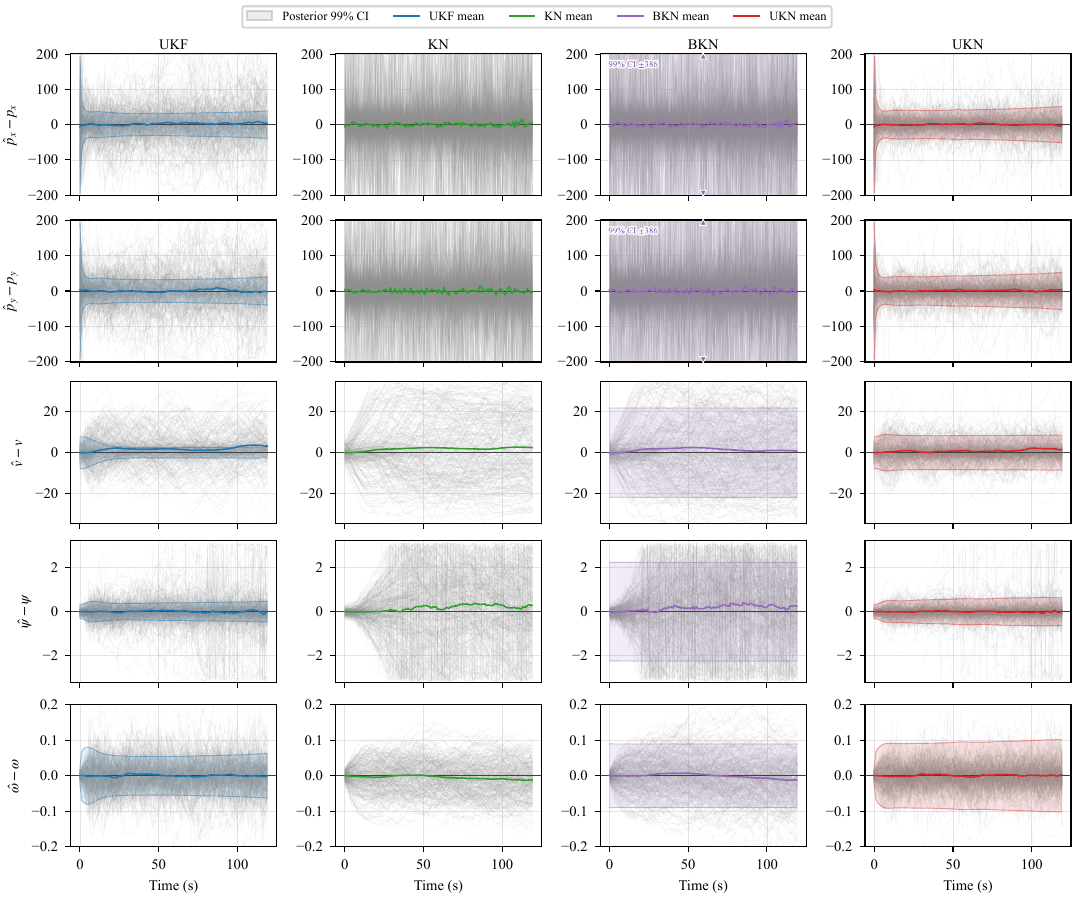}
    \vspace{-10pt}
    \caption{Per-component estimation error $\hat{\bm{x}}_t-\bm{x}_t$ over the test set for the maneuvering target tracking. The gray traces show the realized error for every test trajectory. The KN produces no posterior covariance and therefore no interval.}
    \vspace{-1em}
  \label{fig:ct3}
\end{figure*}

\begin{figure}[!t]
    \centering
    \includegraphics[width=0.55\linewidth]{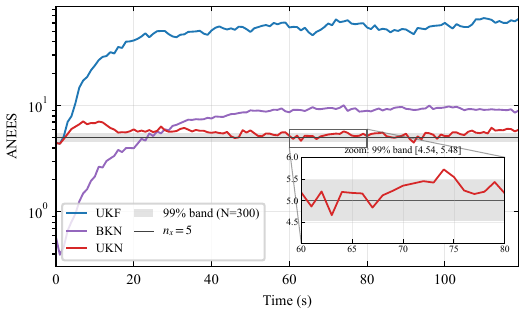}
    \vspace{-5pt}
    \caption{ANEES over the $N=300$ test trajectories for the maneuvering target tracking example. The solid horizontal line denotes the nominal value $n_x=5$, and the horizontal shaded band denotes the pointwise 99\% ANEES consistency interval $[4.542,5.483]$.}
  \label{fig:ct5}
\end{figure}

\subsubsection{Results}
Fig.~\ref{fig:ct1} shows the position estimates for a representative test trajectory, with glint-contaminated measurements marked by orange crosses. The UKF, KN, and BKN exhibit larger deviations from the true path, especially near maneuvering segments and glint-contaminated returns. In contrast, the UKN remains closer to the ground truth, indicating that its learned components mitigate the combined effects of the unmodeled maneuver command and heavy-tailed measurement errors.

Fig.~\ref{fig:ct2} shows the corresponding state components and the marginal 99\% posterior intervals for the covariance-reporting filters. All four filters capture the broad position trend because the radar measurement directly constrains the position through range and bearing. However, their behavior differs substantially in the indirectly observed states. The speed and turn-rate estimates by KN and BKN remains nearly flat despite the substantial variation in the true values. In addition, their heading estimates deviate markedly from the ground truth. The KN does not provide posterior intervals, whereas the BKN reports wide intervals for the indirectly observed states in this representative trajectory. In contrast, the UKN tracks the changes in speed, heading, and turn-rate more closely, with its posterior intervals generally encompassing the
ground-truth trajectory.

The normalized RMSE in Fig.~\ref{fig:ct4} confirms this behavior over the test set. Because the state components have different physical units, the RMSE of each component is normalized by the following reference scale:
\begin{equation}
    [1000~\mathrm{m},\,1000~\mathrm{m},\,30~\mathrm{m/s},\,1~\mathrm{rad},\, 0.1~\mathrm{rad/s}]
\end{equation}
The UKN attains the lowest overall error, $0.282$, compared with $0.383$ for the UKF, corresponding to a $26.4\%$ reduction. The KN and BKN are less accurate than the analytical UKF, with overall errors of $0.694$ and $0.714$, respectively. Although the KN and BKN also exhibit larger position errors than the UKF, their performance degradation is most pronounced in the indirectly observed states. For speed, the normalized RMSE increases from $0.287$ for the UKF to $0.516$ for the KN and $0.521$ for the BKN. For heading, it increases from $0.649$ to $1.378$ and $1.377$, respectively. The UKN reduces the speed, heading and turn-rate errors to $0.241$, $0.441$ and $0.378$, respectively, showing that its improvement is largest on the components least directly informed by the radar.

To determine whether the degraded aggregate performance of the learned baselines reflects uniformly poor tracking or a specific sensitivity to glint-contaminated measurements, we separate the position error over clean and glint-contaminated steps. After excluding the first four initialization steps, let $\mathcal{I}_{\mathrm{c}}$ and $\mathcal{I}_{\mathrm{g}}$ denote the test samples satisfying $g_t^{(k)}=0$ and $g_t^{(k)}=1$, respectively, where $k$ indexes the test trajectory. For $a\in\{\mathrm{c},\mathrm{g}\}$, the subset position RMSE and the glint-to-clean RMSE ratio are defined as:
\begin{align}
    \operatorname{RMSE}^{p}_{a}
    &=
    \sqrt{
        \frac{1}{|\mathcal{I}_{a}|}
        \sum_{(k,t)\in\mathcal{I}_{a}}
        \left\|
            \hat{\bm{p}}^{(k)}_{t|t}
            -
            \bm{p}^{(k)}_t
        \right\|_2^2
    },
    \label{eq:ct_subset_rmse}\\
    \Gamma_{\mathrm{g}}
    &=
    \frac{
        \operatorname{RMSE}^{p}_{\mathrm{g}}
    }{
        \operatorname{RMSE}^{p}_{\mathrm{c}}
    },
    \label{eq:ct_glint_ratio}
\end{align}
where $\bm{p}_t=[p_{x,t},p_{y,t}]^\top$.
Direct inversion of the radar measurement is included as a measurement-only diagnostic:
\begin{equation}
    \widetilde{\bm{p}}_t
    =
    \begin{bmatrix}
        r_t\cos\theta_t\\
        r_t\sin\theta_t
    \end{bmatrix}.
    \label{eq:ct_radar_inversion}
\end{equation}
This diagnostic performs neither state prediction nor temporal filtering. For direct radar inversion, $\hat{\bm{p}}_{t|t}$ in Eq.~\eqref{eq:ct_subset_rmse} is replaced by $\widetilde{\bm{p}}_t$. The adherence of each posterior estimate to the current radar return is quantified by the mean absolute post-fit range residual,
\begin{equation}
    \overline{\varepsilon}_{r,\mathrm{post}}
    =
    \frac{1}{|\mathcal{I}|}
    \sum_{(k,t)\in\mathcal{I}}
    \left|
        r_t^{(k)}
        -
        \left\|
            \hat{\bm{p}}^{(k)}_{t|t}
        \right\|_2
    \right|.
    \label{eq:ct_postfit_range}
\end{equation}
A smaller post-fit range residual indicates closer adherence to the instantaneous radar return, but does not by itself indicate a more accurate state estimate. The resulting aggregate statistics are reported in Table~\ref{tab:ct_glint}.

\begin{table}[!t]
\centering
\caption{Post-fit range residual and position RMSE separated
over clean and glint-contaminated measurement steps in the
maneuvering target tracking example.}
\label{tab:ct_glint}
\begin{tabular}{lccccc}
\hline
Method
& \begin{tabular}[c]{@{}c@{}}Post-fit range\\residual (m)\end{tabular}
& \multicolumn{3}{c}{Position RMSE (m)}
& \begin{tabular}[c]{@{}c@{}}RMSE ratio $\Gamma_{\mathrm{g}}$\\
(Glint/Clean) \end{tabular}
\\
\cline{3-5}
& & All & Clean & Glint & \\
\hline
Radar inversion & 0.00  & 122.02 & 51.78 & 354.71 & 6.85 \\
\hline
UKF             & 25.65 & 92.31  & 89.52 & 114.64 & 1.28 \\
KN              & 4.46  & 122.51 & 52.82 & 355.04 & 6.72 \\
BKN             & 4.44  & 122.54 & 52.84 & 355.11 & 6.72 \\
UKN             & 21.03 & 76.23  & 74.17 & 92.90  & 1.25 \\
\hline
\end{tabular}
\end{table}

Table~\ref{tab:ct_glint} shows that the poor aggregate performance of the KN and BKN does not result from uniformly poor tracking. On clean measurement steps, their position RMSEs are $52.82$~m and $52.84$~m, respectively, substantially lower than the UKF value of $89.52$~m. Thus, both learned baselines effectively exploit nominal radar measurements and can outperform the analytical filter when the measurements are not contaminated by glint.

Their behavior changes sharply under glint contamination. The KN and BKN have mean post-fit range residuals of only $4.46$~m and $4.44$~m, respectively, indicating that their posterior position estimates remain close to the instantaneous radar return. Their glint-step RMSEs increase to approximately $355$~m, nearly matching the $354.71$~m error of direct radar inversion. In contrast, the UKF and UKN maintain larger post-fit range residuals and limit the glint sensitivity ratios $\Gamma_{\mathrm{g}}$ to $1.28$ and $1.25$, respectively. The UKN achieves the lowest overall position RMSE, the lowest glint-step RMSE, and a lower clean-step RMSE than the UKF. This robustness is also consistent with the speed, heading, and turn-rate estimates in Fig.~\ref{fig:ct2}. Because these components are not measured directly and must instead be inferred from the temporal evolution of the estimated position, large measurement-driven position errors can propagate into the indirectly observed states. The UKN provides a more balanced response, retaining sensitivity to nominal measurements while attenuating the effect of impulsive radar errors.

Fig.~\ref{fig:ct3} compares the realized errors with the marginal posterior intervals. The UKF intervals are too narrow relative to the realized error on most components, consistent with underestimated process noise, unmodeled maneuver commands, and unmodeled glint. The BKN intervals are excessively wide for the position, speed, and heading components, while its turn-rate error develops a nonzero bias in the second half of the sequence. The UKN intervals more closely match the empirical spread and maintain mean errors near zero across the state components.

Using the same ANEES criterion as in the preceding examples, Fig.~\ref{fig:ct5} evaluates full-state covariance calibration over the $N=300$ independent test trajectories. Since $n_x=5$, the nominal ANEES is five, and the pointwise 99\% consistency interval is $[4.542,5.483]$. The UKF exceeds the upper limit early and remains far above it, indicating substantial underestimation of the realized estimation error. The BKN initially lies below the lower limit, reflecting conservative covariance estimates, but increases over time and subsequently remains above the upper limit. Its covariance scaling therefore changes from conservatism to overconfidence. The UKN produces the ANEES closest to the nominal value and remains within or near the consistency region for much of the sequence. Its intermittent upper-limit exceedances indicate mild localized overconfidence, but the magnitude of its miscalibration is substantially smaller than for the UKF and BKN.

This example combines the strongest effects of partial observability, transition-model mismatch, covariance misspecification, and non-Gaussian measurement errors. The degradation of the KN and BKN is concentrated in the indirectly observed states and is consistent with inadequate recovery of the cross-covariance structure required to correct these components. The UKN achieves a lower normalized RMSE than the UKF on every state component while providing the most consistent posterior covariance among the covariance-reporting filters. These results suggest that retaining a sigma-point cross-covariance backbone is particularly beneficial when indirectly observed states must be recovered under simultaneous model and noise mismatch.

\begin{figure}[!t]
    \centering
    \includegraphics[width=0.55\linewidth]{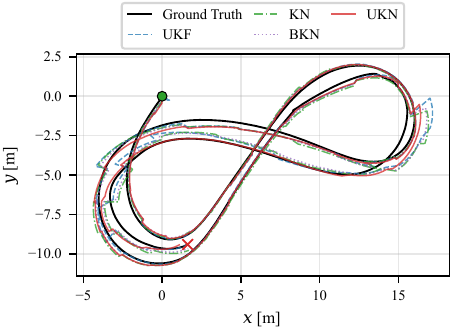}
    \vspace{-10pt}
    \caption{Estimated flight track in the horizontal plane for a representative held-out sequence from the UZH-FPV LOSO evaluation.}
    \vspace{-1em}
  \label{fig:drone1}
\end{figure}

\begin{figure}[!t]
    \centering
    \includegraphics[width=\linewidth]{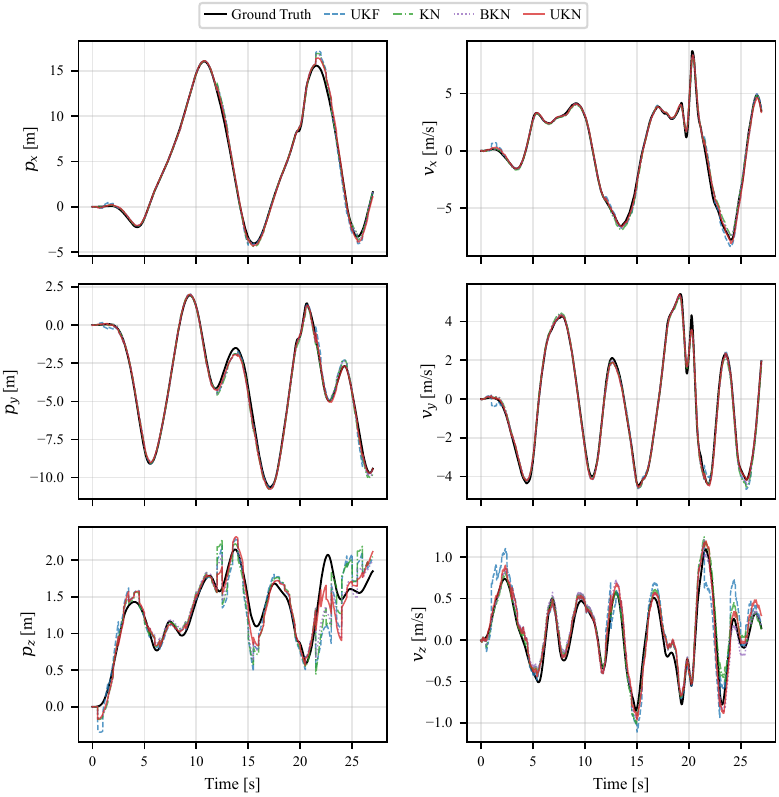}
    \vspace{-10pt}
    \caption{Estimated position and velocity components for the representative LOSO fold shown in Fig.~\ref{fig:drone1}.}
    \vspace{-1em}
  \label{fig:drone2}
\end{figure}

\subsection{Real Flight Data: UZH-FPV Drone Racing}
\label{sec:exp_uzh_fpv}
The final example evaluates the UKN on real flight data from the UZH-FPV indoor drone-racing dataset~\cite{delmerico2019we}. Unlike the previous synthetic examples, the sources of model mismatch and noise misspecification are not prescribed. The filters use onboard IMU acceleration for state propagation and IMU-derived pseudo-velocity observations for measurement updates. Leica ground-truth positions are used only for training and evaluation. This example therefore tests whether the learned correction transfers to a real system in which the mismatch is uncharacterized rather than deliberately constructed.

\subsubsection{Nominal state-space model}
The state $\bm{x}_t=[\bm{p}_t^{\!\top},\dot{\bm{p}}_t^{\!\top}]^{\!\top}
\in\mathbb{R}^6$ consists of the three-dimensional position $\bm{p}_t=[x_{1,t},x_{2,t},x_{3,t}]^{\!\top}$ and velocity $\dot{\bm{p}}_t=[\dot{x}_{1,t},\dot{x}_{2,t},\dot{x}_{3,t}]^{\!\top}$. The nominal transition follows double-integrator kinematics driven by the measured IMU acceleration:
\begin{equation}
\bm{f}(\bm{x}_{t-1},\bm{u}_t)
=
\begin{bmatrix}
\bm{I}_3 & \Delta t\,\bm{I}_3 \\
\bm{0} & \bm{I}_3
\end{bmatrix}
\bm{x}_{t-1}
+
\begin{bmatrix}
\tfrac{1}{2}\Delta t^2\,\bm{I}_3 \\
\Delta t\,\bm{I}_3
\end{bmatrix}
\bm{u}_t,
\label{eq:uzh_f}
\end{equation}
where $\Delta t=0.01$~s and
$\bm{u}_t\in\mathbb{R}^3$ is the onboard IMU linear acceleration used as the propagation input. The filter model is
\begin{equation}
    \bm{x}_t=\bm{f}(\bm{x}_{t-1},\bm{u}_t)+\bm{w}_t .
\end{equation}

The pseudo-measurement model observes the velocity component of the state,
\begin{equation}
\bm{y}_t = \bm{h}(\bm{x}_t)+\bm{v}_t, \qquad \bm{h}(\bm{x}_t) =
    \begin{bmatrix}
        \bm{0} & \bm{I}_3
    \end{bmatrix}
\bm{x}_t .
\label{eq:uzh_h}
\end{equation}
The measurement is an IMU-derived pseudo-velocity rather than a directly measured velocity. No position measurements are provided to the filters during the measurement updates.

\subsubsection{Pseudo-velocity observations}
\label{sec:pseudo_velocity}
The synchronized data are processed on a $100$~Hz time grid, and each sequence is truncated to $T=2700$ steps, corresponding to approximately $27$~s. The state propagation is performed at the full $100$~Hz rate using the onboard IMU linear acceleration. For the measurement update, pseudo-velocity observations are constructed by numerically integrating the IMU acceleration. These observations are supplied to the filters once every fifty steps, corresponding to an effective update rate of $2$~Hz. The intervening steps are treated as missing-measurement steps.

The pseudo-velocity is not an externally measured ground-truth velocity. Its error includes IMU noise and bias, numerical integration drift, and errors arising from the discrepancy between the nominal double-integrator model and the actual quadrotor motion. Because the same IMU acceleration signal is used for state propagation and for constructing the pseudo-velocity, the observation errors may be temporally correlated and may share common IMU-induced components with the propagation errors. These dependencies are not explicitly represented by the nominal filter.

\subsubsection{Unknown mismatch and nominal covariances}
The filters use the nominal transition in Eq.~\eqref{eq:uzh_f} with a fixed process covariance based on the discrete white-acceleration model~\cite{barshalom2001estimation}:
\begin{equation}
    \bar{\bm Q}_{\mathrm{nom}} =(\sigma_a^{\mathrm{nom}})^2
        \begin{bmatrix}
            \tfrac{\Delta t^4}{4}\bm{I}_3 & \tfrac{\Delta t^3}{2}\bm{I}_3 \\
            \tfrac{\Delta t^3}{2}\bm{I}_3 & \Delta t^2\,\bm{I}_3
        \end{bmatrix}
\end{equation}
where $\sigma_a^{\mathrm{nom}}=0.2$. Because $\bar{\bm Q}_{\mathrm{nom}}$ is positive semidefinite, whereas NoiseNet requires a positive-definite baseline covariance for its Cholesky parameterization, we add an isotropic process-noise floor that also represents residual uncertainty not captured by the nominal double-integrator model:
\begin{equation}
\bm{Q}^{\mathrm{nom}}
=
\bar{\bm Q}_{\mathrm{nom}} + 10^{-6} \bm{I}_6
\label{eq:uzh_q}
\end{equation}
The nominal measurement covariance is interpreted as an effective covariance for the pseudo-velocity observations:
\begin{equation}
\bm{R}^{\mathrm{nom}}
=
(\sigma_v^{\mathrm{nom}})^2\bm{I}_3,
\qquad
\sigma_v^{\mathrm{nom}}=0.1 .
\label{eq:uzh_r}
\end{equation}
The nominal baseline covariances are held fixed throughout each flight, while NoiseNet produces time-varying corrections relative to these baselines.

Unlike the synthetic examples, the true process and observation error statistics are unavailable, so the direction and magnitude of the covariance misspecification cannot be specified. The incomplete physics is also uncharacterized. It includes the discrepancy between the nominal double-integrator kinematics in Eq.~\eqref{eq:uzh_f} and the actual quadrotor dynamics, such as aerodynamic drag, thrust dynamics, actuation delay, attitude-dependent acceleration errors, and other unmodeled effects. 

\begin{figure*}[!t]
    \centering
    \includegraphics[width=\linewidth]{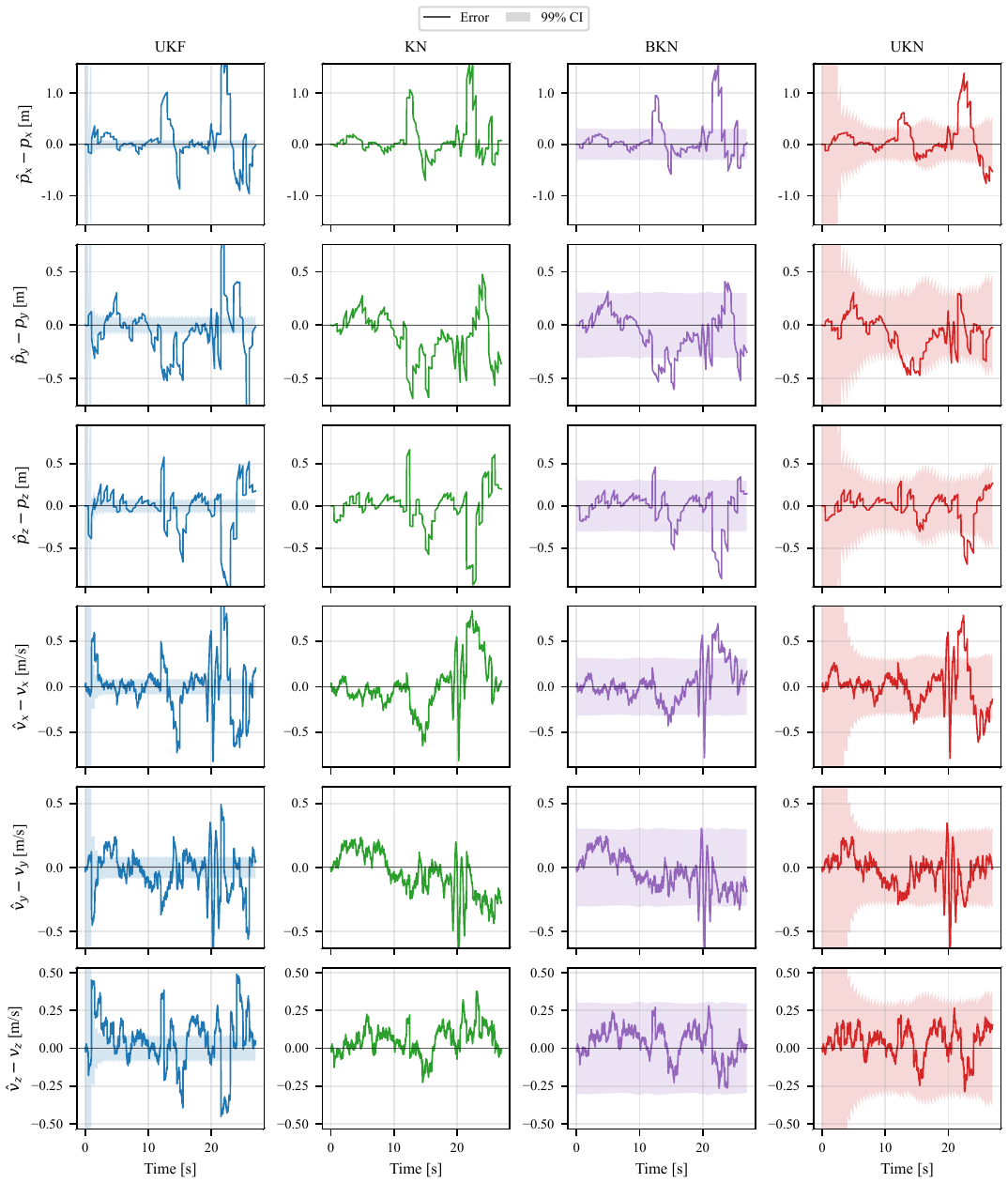}
    \vspace{-10pt}
    \caption{Per-component estimation error for a representative LOSO fold, with the marginal 99\% interval obtained from the diagonal of the posterior covariance. The KN produces no posterior covariance and therefore no interval.}
    \vspace{-1em}
  \label{fig:drone3}
\end{figure*}


\subsubsection{Dataset and evaluation}

The evaluation uses 11 indoor sequences (\texttt{Indoor forward facing \#3, 5, 6, 7, 9, 10} and \texttt{Indoor 45 degree downward facing \#2, 4, 9, 12, 13}) from the UZH-FPV
drone-racing dataset. Generalization across flights is assessed using leave-one-sequence-out (LOSO) cross-validation. In each of the 11 folds, one sequence is withheld exclusively for testing, while the remaining ten sequences are used for training. The reported aggregate metrics are computed from the 11 held-out-sequence results.

The Leica tracking system provides a position reference but no independently measured velocity ground truth is available. Accordingly, the training loss is applied only to the position components of the state. Neither a measured nor a derived velocity reference is used as a training target.


For velocity evaluation, a reference is derived offline from the Leica position trajectory. Specifically, the resampled position is first smoothed using a Savitzky-Golay filter with a 101-sample window, corresponding to approximately 1.0~s on the 100-Hz time grid, and polynomial order of three. A central difference is then applied, and the resulting velocity is smoothed again using the same Savitzky-Golay configuration. The resulting signal is therefore a smoothed derivative-based velocity reference rather than independently measured velocity ground truth.


In each LOSO fold, the Leica position and the derived velocity reference of the held-out sequence are withheld from the filters and used only for evaluation. The pseudo-velocity measurements supplied to the filters are constructed separately by integrating the onboard IMU acceleration, as described in Section~\ref{sec:pseudo_velocity}. All filters use the same training sequences, held-out sequence, nominal covariances, pseudo-velocity observations, initialization, and evaluation references within each fold.

\subsubsection{Results}
Fig.~\ref{fig:drone1} shows the estimated trajectory in the horizontal position plane for the held-out UZH-FPV sequence. All filters capture the overall loop structure, but the UKF and KN exhibit visible offsets along several portions of the path. The BKN reduces part of this deviation, while the UKN remains closest to the Leica reference over most of the sequence.

Fig.~\ref{fig:drone2} compares the position and velocity components over time. The differences among the filters are most visible in the vertical position $p_z$ and velocity components $v_z$. The UKF and KN show larger deviations in these components, whereas the BKN and UKN track the reference more closely. Among them, the UKN provides the closest agreement over the held-out sequence.


\begin{table}[!t]
\centering
\caption{State-estimation accuracy from leave-one-sequence-out cross-validation on the 11 UZH-FPV sequences. Values are reported as the mean $\pm$ standard deviation and range over the 11 held-out sequences.}
\label{tab:drone_loso_accuracy}
\begin{tabular}{llcc}
\hline
Metric & Method & Mean $\pm$ std & Range\\
\hline
Position
& UKF & $0.5488\pm0.0830$ & $[0.4042,0.6850]$ \\
& KN  & $0.5091\pm0.0882$ & $[0.3866,0.6430]$ \\
& BKN & $0.4762\pm0.0744$ & $[0.3533,0.5890]$ \\
& UKN & $0.4261\pm0.0537$
      & $[0.3529,0.5382]$\\
\hline
Velocity
& UKF & $0.4075\pm0.1343$ & $[0.2875,0.6684]$ \\
& KN  & $0.2970\pm0.0555$ & $[0.2254,0.3923]$ \\
& BKN & $0.2851\pm0.0485$ & $[0.2189,0.3638]$ \\
& UKN & $0.2678\pm0.0316$
      & $[0.2041,0.3181]$\\
\hline
\end{tabular}
\end{table}

The aggregate LOSO results in Table~\ref{tab:drone_loso_accuracy} confirm that the observed improvement is not specific to a single flight. The UKN achieves the lowest mean position RMSE, $0.4261\pm0.0537$~m, corresponding to a $22.4\%$ reduction relative to the UKF. The UKN also achieves the lowest mean velocity RMSE, $0.2678\pm0.0316$~m/s, representing a $34.3\%$ reduction relative to the UKF. Moreover, the UKN exhibits the smallest fold-to-fold standard deviation for both metrics, indicating more consistent generalization across flights.


\begin{table}[!t]
\centering
\caption{Covariance-calibration results from LOSO cross-validation. For each fold, the mean dimension-normalized NEES and empirical coverage are computed over the held-out sequence. Values are reported as the mean $\pm$ standard deviation of the fold-wise metrics over the 11 folds.}
\label{tab:drone_loso_calibration}
\begin{tabular}{lccc}
\hline
Method
& \begin{tabular}[c]{@{}c@{}}Mean dimension-normalized \\NEES\end{tabular}
& 95\% coverage
& 99\% coverage \\
\hline
UKF & $20.34\pm7.59$ & $2.8\pm1.0\%$ & $4.4\pm1.7\%$ \\
KN  & -- & -- & -- \\
BKN & $3.31\pm0.78$ & $53.9\pm13.1\%$ & $63.5\pm13.1\%$ \\
UKN & $1.82\pm0.62$
    & $91.2\pm3.7\%$
    & $96.3\pm5.3\%$ \\
\hline
\end{tabular}
\end{table}

Fig.~\ref{fig:drone3} illustrates the state-estimation errors and marginal posterior intervals for the representative LOSO fold. The UKF intervals are narrow relative to the realized errors, whereas the BKN provides wider intervals but still exhibits noticeable drift in several components. The UKN intervals more closely follow the scale of the realized errors, although undercoverage remains over parts of the sequence. This representative behavior is evaluated more systematically across all held-out flights using the LOSO calibration metrics reported in Table~\ref{tab:drone_loso_calibration}.


The LOSO calibration results in Table~\ref{tab:drone_loso_calibration} show that the UKN provides the most reliable covariance behavior across the held-out flights. The NEES is normalized by the state dimension $n_x$ (i.e., $\varepsilon^{\mathrm{N}}_t =\bm e_t^\top\hat{\bm P}_{t|t}^{-1}\bm e_t/n_x$), so that its nominal value is one. Its mean dimension-normalized NEES is $1.82\pm0.62$, compared with $3.31\pm0.78$ for the BKN and $20.34\pm7.59$ for the UKF. Although the UKN value remains above the nominal value of one, it is substantially closer to the nominal consistency level than the other filters. The UKN also achieves empirical coverage of $91.2\pm3.7\%$ and $96.3\pm5.3\%$ for the full-state 95\% and 99\% posterior ellipsoids, respectively. These values are closest to the nominal coverage levels, whereas the BKN exhibits substantial undercoverage and the UKF coverage remains below 5\% at both levels. The LOSO results indicate that the improvements in accuracy and covariance behavior generalize across flights rather than being restricted to a single held-out sequence.


\section{Conclusion}
\label{sec:conclusion}
This paper studied model-based and data-driven information fusion for nonlinear state estimation under incomplete physics with reliable uncertainty quantification. We proposed the Unscented KalmanNet (UKN), a hybrid recursive estimator that augments the Unscented Kalman Filter (UKF) with two learned components while retaining the sigma-point moment propagation and an explicit covariance recursion. NoiseNet replaces the fixed process and measurement covariances with time-varying estimates computed from the filter's internal statistics, and GainNet applies a bounded residual correction to the analytical gain. Both components are anchored to the nominal filter: NoiseNet is initialized to the identity multiplier and GainNet to a zero residual, so the estimator reduces to the UKF at initialization and departs from it only as supported by the data. The two components are trained jointly through a composite objective that combines state-estimation accuracy with covariance calibration and innovation consistency.

The numerical study evaluated the UKN against the UKF, the KN, and the BKN on three synthetic systems and the UZH-FPV real-flight dataset. The UKN attains the lowest aggregate state-estimation error in all four examples and provides the most reliable covariance behavior among the filters that report uncertainty. The KN produces no posterior covariance, so its uncertainty cannot be evaluated. The BKN produces a covariance through sampling, but its uncertainty is often only partially calibrated. The UKF provides an analytical covariance, but with fixed nominal noise statistics, it becomes overconfident in all four examples. In contrast, the UKN improves the state estimates while producing covariance estimates that better match the realized errors. In the synthetic examples, the UKN produces the ANEES closest to the nominal value.

The clean/glint decomposition in the coordinated-turn example further shows that the degraded aggregate performance of the KN and BKN does not result from uniformly poor tracking. Both learned baselines achieve accurate position estimates on clean measurement steps, but their errors under glint contamination approach those of direct radar inversion. The UKN instead achieves the lowest overall and glint-step position RMSE while maintaining a substantially lower sensitivity to contaminated measurements. These results indicate that the UKN provides a more effective balance between responsiveness to nominal measurements and robustness to impulsive measurement errors.

Generalization on the UZH-FPV dataset was assessed using leave-one-sequence-out cross-validation over 11 flight sequences. The UKN reduces the mean position and velocity RMSEs by $22.4\%$ and $34.3\%$, respectively, compared to the UKF, while attaining the lowest mean errors and the smallest fold-to-fold variability among the evaluated methods. It also provides the mean dimension-normalized NEES closest to one and the empirical posterior coverage closest to the nominal $95\%$ and $99\%$ levels among the covariance-reporting filters. These results show that the improvements in accuracy and covariance calibration generalize across flights rather than being restricted to a single held-out sequence.

The present formulation is limited to state estimation, and its componentwise calibration objective directly targets the marginal posterior variances rather than the complete multivariate covariance structure. Future work may therefore consider multivariate calibration objectives involving the full posterior covariance. A further direction is an augmented formulation for joint state and parameter estimation. In such problems, unknown parameters are augmented to the state but are not directly observed and must instead be inferred through their influence on the system dynamics. The sigma-point state-measurement cross-covariance retained as the analytical backbone of the UKN provides a natural mechanism for this inference. Extending the method to augmented estimation will, however, require addressing stronger nonlinearities and distinguishing parameter drift from ordinary process noise. These developments would broaden the present framework from robust state estimation to adaptive joint inference under model uncertainty.



\printcredits

\bibliographystyle{model1-num-names}

\bibliography{cas-refs}




\clearpage
\end{document}